%% file: main.tex
\documentclass[letterpaper]{article} 
\usepackage[submission]{aaai23}  
\usepackage{times}  
\usepackage{helvet}  
\usepackage{courier}  
\usepackage[hyphens]{url}  
\usepackage{graphicx} 
\usepackage{natbib}  
\usepackage{caption} 
\DeclareCaptionStyle{ruled}{labelfont=normalfont,labelsep=colon,strut=off} 
\usepackage[compact]{titlesec}
\titlespacing*{\section}{5pt}{3pt}{2pt}
\titlespacing*{\subsection}{2pt}{2pt}{1pt}

\usepackage{dsfont}
\usepackage{amsmath,amsfonts,amssymb,mathtools, amsthm}
\usepackage[noend]{algorithmic}
\usepackage[ruled,vlined,linesnumbered]{algorithm2e}
\usepackage{tikz, subcaption}
\usepackage[inline]{enumitem}

\usepackage{multirow}
\usepackage{array}
\newcolumntype{M}[1]{>{\centering\arraybackslash}m{#1}}
\newcolumntype{N}{@{}m{0pt}@{}}

\DeclareMathAlphabet\mathbfcal{OMS}{cmsy}{b}{n}

\usetikzlibrary{shapes, arrows, arrows.meta, positioning}

\newcommand{\independent}{\perp\!\!\!\perp}
\newcommand{\notindependent}{\not\!\perp\!\!\!\perp}
\newcommand{\sep}[4]{ (#1 \independent #2 \vert #3)_{#4}}
\newcommand{\nsep}[4]{ (#1 \notindependent #2 \vert #3)_{#4}}

\DeclareMathOperator*{\argmin}{arg\,min}
\DeclareMathOperator*{\argmax}{arg\,max}

\newcommand{\Pa}[2]{\textit{Pa}_{#2}(#1)}
\newcommand{\Ch}[2]{\textit{Ch}_{#2}(#1)}
\newcommand{\Anc}[2]{\textit{Anc}_{#2}(#1)}

\newcommand{\Mb}[2]{\textit{Mb}_{#2}(#1)}

\newcommand{\V}[0]{\mathbf{V}}
\newcommand{\C}[0]{\mathbf{C}}

\newcommand{\U}[0]{\mathbf{U}}
\newcommand{\E}[0]{\mathbf{E}}
\newcommand{\X}[0]{\mathbf{X}}

\newcommand{\Z}[0]{\mathbf{Z}}

\newcommand{\N}[0]{\mathbf{N}}

\newcommand{\G}[0]{\mathcal{G}}

\newcommand{\PV}{P_{\V}}
\newcommand{\Gpi}{\G^{\pi}}
\newcommand{\Data}{\textit{Data}}
\newcommand{\Vrem}{\V_{\text{rem}}}

\newtheorem{theorem}{Theorem}
\newenvironment{customthm}[1]{\renewcommand\thetheorem{#1}\theorem}{\endtheorem}
\newtheorem{corollary}{Corollary}
\newtheorem{lemma}{Lemma}

\newtheorem{proposition}{Proposition}
\newenvironment{customprp}[1]{\renewcommand\theproposition{#1}\proposition}{\endproposition}
\newtheorem{definition}{Definition}
\newtheorem{remark}{Remark}
\newtheorem*{lemma*}{Lemma}

\title{Novel Ordering-based Approaches for Causal Structure Learning in the Presence of Unobserved Variables}
\author{
    Ehsan Mokhtarian,\textsuperscript{\rm 1}
    Mohammadsadegh Khorasani,\textsuperscript{\rm 1}
    Jalal Etesami,\textsuperscript{\rm 1}
    Negar Kiyavash\textsuperscript{\rm 1 2} 
}
\affiliations{
    \textsuperscript{\rm 1}School of Computer and Communication Sciences EPFL, Lausanne, Switzerland\\
    \textsuperscript{\rm 2}College of Management of Technology EPFL, Lausanne, Switzerland\\
    \{ehsan.mokhtarian, sadegh.khorasani, seyed.etesami, negar.kiyavash\}@epfl.ch

}

\begin{document}

\maketitle
\begin{abstract}
    We propose ordering-based approaches for learning the maximal ancestral graph (MAG) of a structural equation model (SEM) up to its Markov equivalence class (MEC) in the presence of unobserved variables. Existing ordering-based methods in the literature recover a graph through learning a causal order (c-order). We advocate for a novel order called removable order (r-order) as they are advantageous over c-orders for structure learning. This is because r-orders are the minimizers of an appropriately defined optimization problem that could be either solved exactly (using a reinforcement learning approach) or approximately (using a hill-climbing search). Moreover, the r-orders (unlike c-orders) are invariant among all the graphs in a MEC and include c-orders as a subset. Given that set of r-orders is often significantly larger than the set of c-orders, it is easier for the optimization problem to find an r-order instead of a c-order. We evaluate the performance and the scalability of our proposed approaches on both real-world and randomly generated networks.
\end{abstract}

\section{Introduction}
    A causal graph is a probabilistic graphical model that represents conditional independencies (CIs) among a set of observed variables $\V$ with a joint distribution $\PV$.
    When all the variables in the system are observed (i.e., causal sufficiency holds), a causal graph is commonly modeled with a directed acyclic graph (DAG), $\G$.
    It is well-known that from mere observational distribution $\PV$, graph $\G$ can only be learned up to its Markov equivalence class (MEC) \citep{spirtes2000causation,pearl2009causality}. 
    Therefore, the problem of causal structure learning (aka causal discovery) from observational distribution in the absence of latent variables refers to identifying the MEC of $\G$ using a finite set of samples from $\PV$ and has important applications in many areas such as biology \citep{sachs2005causal}, advertisements\citep{bottou2013counterfactual}, social science \citep{russo2010causality}, etc.

    \begin{figure}[t]
        \centering
        \begin{subfigure}{0.3 \textwidth}
            \includegraphics[width=\textwidth]{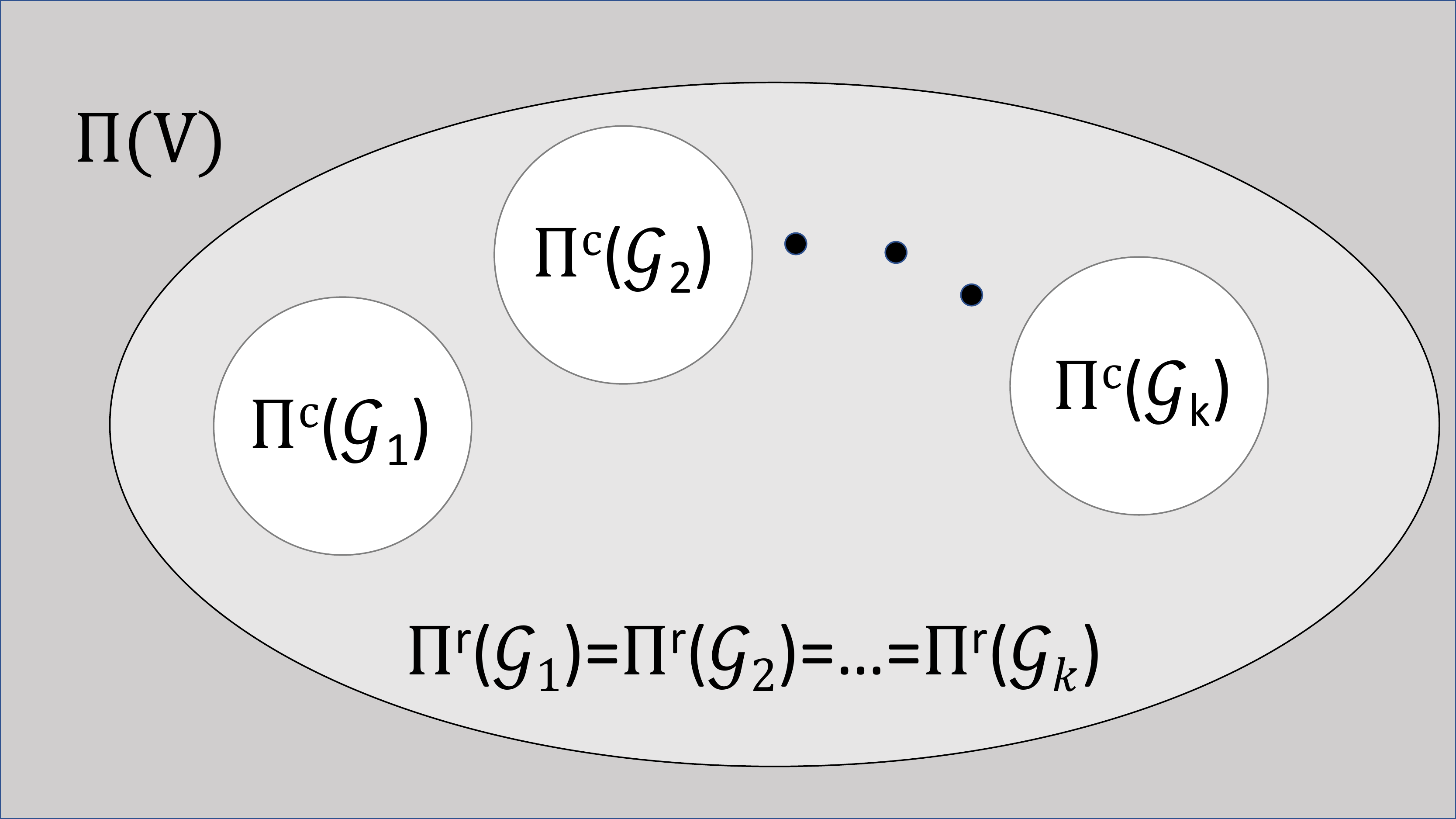}
        \end{subfigure}
        \caption{In this figure, $\{\G_1,\cdots,\G_k\}$ denotes a set of Markov equivalent DAGs.
        $\Pi(\V)$ denotes the set of orders over $\V$, which is the search space of ordering-based methods.
        $\Pi^c(\G_i)$ denotes the set of c-orders of $\G_i$, the target space of existing ordering-based methods in the literature.
        $\Pi^r(\G_i)$ denotes the set of r-orders of $\G_i$, which is the target space of the proposed methods in this paper.}
        \label{fig: venn}
    \end{figure}
    
    There are three main classes of algorithms for causal structure learning: constraint-based, score-based, and hybrid methods.
    Constraint-based methods use the available data from $\PV$ to test for CI relations in the distribution, from which they learn the MEC of $\G$ \citep{zhang2012kernel,spirtes2000causation, colombo2012learning, sun2007kernel,zhang2017learning}.
    Score-based methods define a score function (e.g., regularized likelihood function or Bayesian information criterion (BIC)) over the space of graphs and search for a structure that maximizes the score function \citep{zheng2018dags, zhu2019causal,wang2021ordering}.
    Hybrid methods combine the strength of both constraint-based and score-based methods to improve score-based algorithms by applying constraint-based techniques \citep{tsamardinos2006max}. 
    
    Under causal sufficiency assumption, the search space of most of the score-based algorithms is the space of DAGs, which contain $2^{\Omega(n^2)}$ members when there are $n$ variables in the system.
    In score-based methods, a variety of search strategies are proposed to solve the maximization problem.
    Teyssier and Koller \citep{teyssier2012ordering} introduced the first ordering-based search strategy to solve the score-based optimization.
    The search space of such ordering-based methods is the space of orders over the vertices of the DAG, which includes $2^{\mathcal{O}(n \log(n))}$ orders.
    Note that the space of orders is significantly smaller than the space of DAGs.
    Ordering-based methods such as \citep{zhu2019causal,larranaga1996learning, teyssier2012ordering, friedman2003being}, divide the learning task into two stages.
    In the first stage, they use the available data to find a causal order (c-order) over the set vertices of $\G$. 
    They use the learned order in the second stage to identify the MEC of $\G$.
    
    All the aforementioned ordering-based approaches for causal discovery require causal sufficiency.
    In practice, presence of unobserved variables is more the norm rather than the exception.
    In such cases, instead of a DAG, graphical models such as maximal ancestral graph (MAG) and inducing path graph (IPG) are developed in the literature to represent a causal model \citep{mokhtarian2021recursive, zhang2019recursively, rahman2021accelerating}.
    We introduce a novel type of order for MAGs, called removable order (in short, r-order), and argue that r-orders are advantageous over c-orders for structure learning.
    For one, as r-orders are defined for MAGs (as opposed to DAGs), they can be used to design algorithms for causal graph structure discovery in the absence of causal sufficiency.
    Moreover, even in the absence of latent variables, r-orders are better suited for learning the MEC. 
    This is because, as depicted in Figure \ref{fig: venn}, r-orders include c-orders.
    As a consequence, the problem of searching for an r-order is easier than finding a c-order as the search space remains the same, but the set of feasible solutions is larger.

    Our main contributions are summarized as follows.
    \begin{enumerate}[leftmargin=*]
    
        \item We introduce a novel type of order for MAGs, called r-order, which are invariant among the MAGs in a MEC (Proposition \ref{prp: r-order is learnable}).
        In the case of DAGs, we note that this property does not hold for c-orders as they are mutually exclusive across the graphs in a MEC (Figure \ref{fig: venn}).
        \item We propose ordering-based approaches for identifying the MEC of a MAG using r-orders.
        In particular, our methods do not require causal sufficiency. 
        Furthermore, we show that the problem of finding an r-order can be cast as a minimization problem and prove that r-orders are the unique minimizers of this problem (Theorem \ref{thm: opt prob}).
        \item We show that our minimization problem can be formulated with appropriately defined costs as a reinforcement learning problem. 
        Accordingly, any reinforcement learning algorithm can be applied to find a solution for our problem. 
        Additionally, we propose a hill-climbing search algorithm to approximate the solution of the optimization problem of our interest.
    \end{enumerate}

\subsection{Related Work}
    Under causal sufficiency assumption, several causal structure discovery approaches have been proposed in the literature: constraint-based \citep{spirtes2000causation, margaritis1999bayesian, pellet2008using, mokhtarian2021recursive, tsamardinos2003time,sun2007kernel, mokhtarian2021learning}, score-based \citep{nandy2018high, zheng2018dags, bottou2013counterfactual, yu2019dag}, and hybrid \citep{nandy2018high, gamez2011learning, schulte2010imap, schmidt2007learning,alonso2013scaling}.
    Some score-based methods such as \citep{yu2019dag, lachapelle2019gradient, ng2022masked, zheng2020learning} formulate the structure learning problem as a smooth continuous optimization and exploit gradient descent to solve it.
    In \citep{zhu2019causal, wang2021ordering}, the optimization problem is formulated as a reinforcement learning problem, where the score function is defined over DAGs in  \citep{zhu2019causal} and over orders in  \citep{wang2021ordering}.
    Furthermore, among score-based approaches, various ordering-based methods such as \citep{zhu2019causal, larranaga1996learning, teyssier2012ordering, friedman2003being} have been proposed that exploit different search strategies to find a c-order.
    All of these ordering-based approaches are heuristics and provide no guarantees to finding a correct c-order.
    
    There are a few papers in the literature that do not require causal sufficiency.
    FCI \citep{spirtes2000causation} is a constraint-based algorithm that starts with the skeleton of the graph learned by PC algorithm and then performs more CI tests to learn a MAG up to its MEC.
    RFCI \citep{colombo2012learning}, FCI+ \citep{claassen2013learning}, and MBCS*\citep{pellet2008finding} are three modifications of FCI.
    L-MARVEL \citep{akbari2021recursive} is a recursive algorithm that iteratively eliminates specific variables and learns the skeleton of a MAG.
    M3HC \cite{tsirlis2018scoring} is a hybrid method that can learn a MAG up to its MEC.
    To the best of our knowledge, the only other work in the literature that uses an ordering-based approach for causal discovery in MAGs (i.e., in the presence of latent variable) is GSPo which proposes a greedy algorithm that is only consistent as long as there are no latent variables in the system (the graph is a DAG) \citep{raskutti2018learning}, but there are no theoretical guarantees in case of MAGs \citep{bernstein2020ordering}.

\section{Preliminary and Problem Description}
    Throughout the paper, we denote random variables by capital letters (e.g., $X$) and sets of variables by bold letters (e.g., $\X$).
    A \emph{mixed graph} (MG) is a graph $\G = (\V,\E_1,\E_2)$, where $\V$ is a set of vertices, $\E_1$ is a set of directed edges, i.e., $\E_1 \subseteq \{(X,Y)\mid X,Y \in \V \}$, and $\E_2$ is a set of bidirected edges, i.e., $\E_2 \subseteq \{\{X,Y\}\mid X,Y \in \V \}$.
    For a subset $\Z \subseteq \V$, MG $\G[\Z]= (\Z,\E_1^{\Z},\E_2^{\Z})$ denotes the induced subgraph of $\G$ over $\Z$, that is $\E_1^{\Z} = \{(X,Y)\in \E_1 \mid X,Y \in \Z \}$ and $\E_2^{\Z} = \{\{X,Y\}\in \E_2 \mid X,Y \in \Z \}$.
    For each directed edge $(X,Y)$ in $\E_1$, we say $X$ is a \emph{parent} of $Y$ and $Y$ is a \emph{child} of $X$.
    Further, we say $X$ and $Y$ are neighbors if a directed or undirected edge exists between them in $\G$.
    The \emph{skeleton} of $\G$ is the undirected graph obtained by removing the directions of the edges of $\G$.
    A path $(X_1,X_2,\cdots\!,X_k)$ in $\G$ is called a \emph{directed path} from $X_1$ to $X_k$ if $(X_i,X_{i+1})\in \E_1$ for all $1\leq i<k$. 
    If a directed path exists from $X$ to $Y$, $X$ is called an \emph{ancestor} of $Y$.
    We denote the set of parents, children, and ancestors of $X$ in $\G$ by $\Pa{X}{\G}$, $\Ch{X}{\G}$, and $\Anc{X}{\G}$, respectively.
    We also apply these definitions disjunctively to sets of variables, e.g., $\Anc{\X}{\G} = \bigcup_{X\in \X}\Anc{X}{\G}$.
    A non-endpoint vertex $X_i$ on a path $(X_1,X_2, \cdots,X_k)$ is called a \emph{collider}, if one of the following situations arises.
    \begin{align*}
        X_{i-1} \to X_i \gets X_{i+1},\quad X_{i-1} \leftrightarrow X_i \gets X_{i+1},\\ 
        X_{i-1} \to X_i \leftrightarrow X_{i+1}, \quad
        X_{i-1} \leftrightarrow X_i \leftrightarrow X_{i+1}.
    \end{align*}
    A path $\mathcal{P}= (X, W_1, \cdots, W_k, Y)$ between two distinct variables $X$ and $Y$ is said to be \emph{blocked} by a set $\Z \subseteq \V \setminus \{X,Y\}$ in $\G$ if there exists $1 \leq i \leq k$ such that 
    \begin{enumerate*}[label=(\roman*)]
        \item $W_i$ is a collider on $\mathcal{P}$ and $W_i \notin \Anc{\Z \cup \{X,Y\}}{\G}$, or
        \item $W_i$ is not a collider on $\mathcal{P}$ and $W_i \in \Z$. 
    \end{enumerate*}
    We say $\Z$ \emph{m-separates} $X$ and $Y$ in $\G$ and denote it by $\sep{X}{Y}{\Z}{\G}$ if all the paths in $\G$ between $X$ and $Y$ are blocked by $\Z$.
    
    A \emph{directed cycle} exists in an MG $\G=(\V,\E_1,\E_2)$ when there exists $X,Y \in \V$ such that $(X,Y) \in \E_1$ and $Y\in \Anc{X}{\G}$.
    Similarly, an \emph{almost directed cycle} exists in $\G$ when there exists $X,Y \in \V$ such that $\{X,Y\} \in \E_2$ and $Y\in \Anc{X}{\G}$.
    An MG with no directed cycles or almost-directed cycles is said to be \emph{ancestral}.
    An ancestral MG is called \emph{maximal} if every pair of non-neighbor vertices are m-separable, i.e., there exists a set of vertices that m-separates them.
    An MG is called a \emph{maximal ancestral graph} (MAG) if it is both ancestral and maximal.
    A MAG with no bidirected edges is called a \emph{directed acyclic graph (DAG)}.
    Two MAGs $\G_1$ and $\G_2$ are \emph{Markov equivalent} if they impose the same set of m-separations, i.e., $\sep{X}{Y}{\Z}{\G_1} \iff \sep{X}{Y}{\Z}{\G_2}$.
    We denote by $[\G]$ the Markov equivalence class (MEC) of MAG $\G$, i.e., the set of Markov equivalent MAGs of $\G$.
    Moreover, if $\G$ is a DAG, we denote by $[\G]^{\text{DAG}}$ the set of Markov equivalent DAGs of $\G$.

    Let $\G=(\V,\E_1,\E_2)$ and $\PV$ denote a MAG and a joint distribution over set of vertices $\V$, respectively.
    For two distinct variables $X$ and $Y$ in $\V$ and a subset $\Z \subseteq \V \setminus \{X,Y\}$, if $P(X,Y \mid \Z) = P(X \mid \Z) P(Y \mid \Z)$, then $X$ and $Y$ are said be conditionally independent given $\Z$ and it is denoted by $\sep{X}{Y}{\Z}{\PV}$.
    A Conditional Independence (CI) test refers to detecting whether $\sep{X}{Y}{\Z}{\PV}$.
    MAG $\G$ satisfies \emph{faithfulness} w.r.t. (is faithful to) $\PV$ if m-separations in $\G$ is equivalent to CIs in $\PV$, i.e., 
    \begin{equation*}
        \sep{X}{Y}{\Z}{\G} \iff \sep{X}{Y}{\Z}{\PV}.    
    \end{equation*}
    
    

    \subsection{Problem Description:}
        Consider a set of variables $\V \cup \U$, where $\V$ and $\U$ denote the set of observed and unobserved variables, respectively.
        In a \emph{structural equation model} (SEM), each variable $X \in \V\cup \U$ is generated as $X = f_X(\Pa{X}{},\epsilon_X)$, where $f_X$ is a deterministic function,  $\Pa{X}{} \subseteq \V\cup \U \setminus \{X\}$, and $\epsilon_X$ is the exogenous variable corresponding to $X$ with an additional assumption that the exogenous variables are jointly independent \citep{pearl2009causality}.
        The causal graph of an acyclic SEM is a directed acyclic graph (DAG) over $\V \cup \U$ obtained by adding a directed edge from each variable in $\Pa{X}{}$ to $X$, for  $X\in \V \cup \U$.
        The \emph{latent projection} of this DAG over $\V$ is a MAG over $\V$ such that for  $X, Y \in \V$ and $\Z \subseteq \V \setminus \{X,Y\}$, we have 
        \begin{equation*}
            \sep{X}{Y}{\Z}{\text{DAG}} \iff \sep{X}{Y}{\Z}{\text{MAG}}.
        \end{equation*}
        For more details regarding the latent projection, please refer to \citep{verma1991equivalence, akbari2021recursive}.
        Let us denote by $\G$ the resulting MAG over $\V$.
        In this paper, we assume faithfulness, i.e., 
        \begin{equation*}
            \sep{X}{Y}{\Z}{\PV} \iff \sep{X}{Y}{\Z}{\G}.
        \end{equation*}
        The assumption of causal sufficiency refers to assuming that $\U = \varnothing$.
        Note that with causal sufficiency, $\G$ is a DAG.
        
        The problem of causal discovery refers to identifying the MEC of $\G$ using samples from the observational distribution.
        We propose three methods for identifying MEC $[\G]$ using a finite set of samples from $\PV$.
        It is noteworthy that our proposed methods do not require causal sufficiency.

\section{Ordering-based Methods: Removable Orders vs Causal Orders}
\begin{figure}[t] 
	    \centering
		\tikzstyle{block} = [circle, inner sep=1.3pt, fill=black]
		\tikzstyle{input} = [coordinate]
		\tikzstyle{output} = [coordinate]
		\begin{subfigure}{0.23\textwidth}
    		\centering
            \begin{tikzpicture}
                \tikzset{edge/.style = {->,> = latex',-{Latex[width=1.5mm]}}}
                \node[block] (X1) at  (1.5,-0.6) {};
                \node[] ()[below=0 of X1]{$X_1$};
                \node[block] (X2) at  (1.5,0.6) {};
                \node[] ()[above=0 of X2]{$X_2$};
                \node[block] (X3) at  (0,0) {};
                \node[] ()[below=0 of X3]{$X_3$};
                \node[block] (X4) at  (3,0) {};
                \node[] ()[below=0 of X4]{$X_4$};
                \draw[edge] (X3) to (X2);
                \draw[edge] (X3) to (X1);
                \draw[edge] (X4) to (X2);
                \draw[edge] (X4) to (X1);
                \draw[edge] (X2) to (X1);
            \end{tikzpicture}
            \captionsetup{justification=centering}
            \caption{DAG $\G_1$,
            \\ $\Pi^c(\G_1) = \{(X_1,X_2,X_3,X_4),$\, 
            \\ $(X_1,X_2,X_4,X_3)\}$.}
            \label{fig: 1a}
        \end{subfigure}
        \hfill
        \begin{subfigure}{0.23\textwidth}
    		\centering
            \begin{tikzpicture}
                \tikzset{edge/.style = {->,> = latex',-{Latex[width=1.5mm]}}}
                \node[block] (X1) at  (1.5,-0.6) {};
                \node[] ()[below=0 of X1]{$X_1$};
                \node[block] (X2) at  (1.5,0.6) {};
                \node[] ()[above=0 of X2]{$X_2$};
                \node[block] (X3) at  (0,0) {};
                \node[] ()[below=0 of X3]{$X_3$};
                \node[block] (X4) at  (3,0) {};
                \node[] ()[below=0 of X4]{$X_4$};
                \draw[edge] (X3) to (X2);
                \draw[edge] (X3) to (X1);
                \draw[edge] (X4) to (X2);
                \draw[edge] (X4) to (X1);
                \draw[edge] (X1) to (X2);
            \end{tikzpicture}
            \captionsetup{justification=centering}
            \caption{DAG $\G_2$,
             \\ $\Pi^c(\G_2) = \{(X_2,X_1,X_3,X_4),\,$ 
             \\ $(X_2,X_1,X_4,X_3)\}$.}
            \label{fig: 1b}
        \end{subfigure}
        \caption{Two Markov equivalent DAGs $\G_1$ and $\G_2$ that form a MEC together and their disjoint sets of c-orders.
        In this example, any order over $\V = \{X_1,X_2,X_3,X_4\}$ is an r-order, i.e., $\Pi^r(\G_1) = \Pi^r(\G_2) = \Pi(\V)$. Note that $|\Pi^r(\G_1)| =|\Pi^r(\G_2)|=24 > 2 = |\Pi^c(\G_1)| =|\Pi^c(\G_2)|$.}
        \label{fig: example}
    \end{figure}
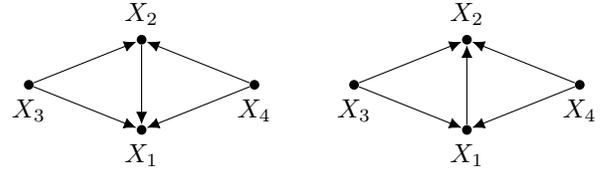

    In this section, we first define \emph{orders} and \emph{c-orders}.
    Then, we introduce our novel order, \emph{r-order}, and provide some of its appealing properties.
    \begin{definition}[order] \label{def: order}
        An $n$-tuple $(X_1,\cdots,X_n)$ is called an \emph{order} over a set $\V$ if $|\V|=n$ and $\V=\{X_1,\cdots, X_n\}$.
        We denote by $\Pi(\V)$, the set of all orders over $\V$.
    \end{definition}
    \begin{definition}[c-order] \label{def: c-order}
        An order $(X_1,\!\cdots\!, X_n) \in \Pi(\V)$ is called a causal order (in short c-order\footnote{Note that this definition is in the opposite direction than usually c-order is defined in the literature.}) of a DAG $\G \!=\! (\V,\E_1,\varnothing)$ if $i\! >\!j$ for each $(X_i, X_j) \in \E_1$. 
        We denote by $\Pi^c(\G)$ the set of c-orders of $\G$.
    \end{definition}
    Consider the two DAGs $\G_1$ and $\G_2$ and their sets of c-orders depicted in Figure \ref{fig: example}.
    In this case, $\G_1$ and $\G_2$ are Markov equivalent and together form a MEC.
    Furthermore, $\Pi^c(\G_1)$ and $\Pi^c(\G_2)$ are disjoint and each contain 2 orders.

    As mentioned earlier, nearly all existing ordering-based methods assume causal sufficiency.
    These methods divide the learning task into two stages.
    In the first stage, they search in $\Pi(\V)$ (which we refer to as the \emph{search space}) to find an order in $\Pi^c(\G)$ (which we refer to as the \emph{target space}).
    In the second stage, they use the discovered order to identify MEC $[\G]^{\text{DAG}}$.
    
    Next, we show that different DAGs in a MEC have a disjoint set of c-orders. 
    In other words, c-orders are not invariant among the DAGs in a MEC.
    Note that detailed proofs appear in the appendix.
    \begin{proposition} \label{prp: not learnable}
        Let $\G$ denotes a DAG with MEC $[\G]^{\text{DAG}} = \{\G_1,...,\G_k\}$. For any two distinct DAGs $\G_i$ and $\G_j$ in $[\G]^{\text{DAG}}$, we have $\Pi^c(\G_i) \cap \Pi^c(\G_j) = \varnothing$.
    \end{proposition}
    \subsection{Removable Orders}
        In this section, we propose a novel set of orders over the vertices of a MAG, called \emph{removable order (in short r-order)}, and show that r-orders are advantageous for structure learning.
        First, we review the notion of a removable variable in a MAG, which was recently proposed in the structure learning literature \citep{mokhtarian2021recursive, mokhtarian2021learning, akbari2021recursive}. 
        
        \begin{definition}[removable variable]\label{def: removable}
            Suppose $\G = (\V,\E_1,\E_2)$ is a MAG.
            A variable $X\in \V$ is called removable in $\G$ if $\G$ and $\G[\V']$ impose the same set of m-separation relations over $\V'$, where $\V' = \V \setminus \{X\}$. 
            That is, for any variables $Y,W\in \V'$ and $\Z \subseteq \V' \setminus \{Y,W\}$,
            \begin{equation*}
                \sep{Y}{W}{\Z}{\G}
    		    \iff
    		    \sep{Y}{W}{\Z}{\G[\V']}.
    		\end{equation*}
        \end{definition}
        Below, we introduce the notion of \emph{r-order}.
        \begin{definition}[r-order] \label{def: r-order}
            An order $\pi = (X_1,\cdots, X_n)$ over set $\V$ is called a removable order (r-order) of a MAG $\G =(\V,\E_1,\E_2)$ if $X_i$ is a removable variable in $\G[\{X_i,\cdots , X_n\}]$ for each $1 \leq i \leq n$. 
            We denote by $\Pi^r(\G)$ the set of r-orders of $\G$. 
        \end{definition}
        Back to the example in Figure \ref{fig: example} where $\G_1$ and $\G_2$ are two Markov equivalent DAGs. 
        In this case, any order over the set of vertices is an r-order for both $\G_1$ and $\G_2$.
        Hence, each graph has 24 r-orders.
        
        In general, all MAGs in a MEC have the same set of r-orders.
        Furthermore, in DAGs, r-orders include all the c-orders as subsets (See Figure \ref{fig: venn}). 
        The following propositions formalize these assertions.
        \begin{proposition} \label{prp: r-order is learnable}
            If $\G_1$ and $\G_2$ are two Markov equivalent MAGs, then $\Pi^r(\G_1) = \Pi^r(\G_2)$.
        \end{proposition}
        \begin{proposition} \label{prp: c-o subset of r-o}
            For any DAG $\G$, we have $\Pi^c(\G) \subseteq \Pi^r(\G)$.
        \end{proposition}
        In light of the above propositions, we can summarize some clear advantages of r-orders as follows:
         
        (i) Implication of Proposition \ref{prp: r-order is learnable} is that, unlike c-ordering-based methods, which fail to find a c-order consistent with all the DAGs within a MEC (Proposition \ref{prp: not learnable}), r-ordering-based methods can find an order which is an r-order for all the MAGs in its corresponding MEC.
        
        (ii) Proposition \ref{prp: c-o subset of r-o} implies that in DAGs, the space of r-orders is in general bigger than the space of c-orders.
        Hence, the target space of an r-ordering-based method is larger than the target space of a c-ordering-based method.
        For instance, in Figure \ref{fig: example}, a c-ordering-based method must find one of the two c-orders of either $\G_1$ or $\G_2$, while an r-ordering-based method can find any of the 24 r-orders in $\Pi^r(\G_1)=\Pi^r(\G_2)$. 
        
        (iii) Since r-orders are defined for MAGs (instead of DAGs), they could be used in ordering-based structure learning approaches without requiring causal sufficiency.
        
\section{Learning an R-order}
    In this section, we describe our approach for learning an r-order of the MAGs in $[\G]$.
    Recall that all MAGs in $[\G]$ have the same set of r-orders.
    We first propose an algorithm that constructs an undirected graph $\Gpi$ corresponding to an arbitrary order $\pi \in \Pi(\V)$.
    Subsequently, we assign a cost to an order $\pi$ based on the constructed graph $\Gpi$, which is simply the number of edges in $\Gpi$, and show that finding an r-order for $[\G]$ can be cast as an optimization problem with the aforementioned cost.
    Then, we propose three algorithms to solve the optimization problem.
    
    \begin{algorithm}[t]
        \caption{Learning $\Gpi$.}
        \label{alg: Gpi}
        \begin{algorithmic}[1]
            \STATE \textbf{Function LearnGPi} ($\pi,\, \Data(\V)$)
            \STATE $\V_1 \gets \V$,  $\E^{\pi}\gets \varnothing$
            \FOR{$t=1$ to $|\V|-1$}
                \STATE $X_t \gets \pi(t)$
                \STATE $\N_{X_t} \gets \textbf{FindNeighbors}(X_t,\, \Data(\V_t))$
                \STATE Add undirected edges between $X_t$ and the variables in $\N_{X_t}$ to $\E^{\pi}$.
                \STATE $\V_{t+1} \gets \V_t \setminus \{X\}$
            \ENDFOR
            \STATE \textbf{Return} $\Gpi = (\V, \E^{\pi})$
        \end{algorithmic}
    \end{algorithm}
    
    \subsection{Learning an Undirected Graph From an Order}
        Algorithm \ref{alg: Gpi} iteratively constructs an undirected graph $\Gpi = (\V,\E^{\pi})$ from a given order $\pi \in \Pi(\V)$.
        The inputs of Algorithm \ref{alg: Gpi} are an order $\pi$ over $\V$ and observational data $\Data(\V)$ sampled from a joint distribution $\PV$.
        The algorithm initializes $\V_1$ with $\V$ and $\E^{\pi}$ with the empty set in lines 2 and 3, respectively. 
        Then in lines 4-8, it iteratively selects a variable $X_t$ according to the given order $\pi$ (line 5) and calls function \textit{FindNeighbors} in line 6 to learn a set $\N_{X_t} \subseteq \V_t \setminus \{X_t\}$.
        Then, the algorithm adds undirected edges to $\Gpi$ to connect $X_t$ and its discovered neighbors $\N_{X_t}$ (line 7).
        Finally, it updates $\V_{t+1}$ by removing $X_t$ from $\V_t$ (line 8) and repeats the process.
        
        The output of function \textit{FindNeighbors}, i.e., $\N_{X_t}$, is the set of variables in $\V_t$ that are not m-separable from $X_t$ using the variables in $\V_t$.
        Hence, if MAG $\G[\V_t]$ is faithful to $P_{\V_t}$, then $N_{X_t}$ would be the set of neighbors of $X_t$ among the variables in $\V_t$.\footnote{Note that non-neighbor variables in any MAG are m-separable.}
        However, since $\pi$ is arbitrary, $\G[\V_t]$ is not necessarily faithful to $P_{\V_t}$ and therefore, $\N_{X_t}$ can include some vertices that are not neighbors of $X_t$.
        There exist several constraint-based algorithms in the literature, such as \citep{spirtes2000causation, pellet2008finding, colombo2012learning, akbari2021recursive} that are designed to verify whether two given variables are m-separable.
        Accordingly, \textit{FindNeighbors} can use any of such algorithms.
        Please note that unlike the methods in \citep{mokhtarian2021learning, akbari2021recursive} where removable variables are discovered in each iteration, Algorithm \ref{alg: Gpi} selects variables according to the given order $\pi$ (line 4).
    
    \subsection{Cost of an Order}
        Suppose $\G$ is faithful to $\PV$.
        It is shown in \citep{akbari2021recursive} that omitting a removable variable does not violate faithfulness in the remaining graph.
        Hence, due to the definition of r-order, if $\pi \in \Pi^r(\G$), then after each iteration $t$, MAG $\G[\V_t]$ remains faithful to $P_{\V_t}$.
        The next result shows that Algorithm \ref{alg: Gpi} constructs the skeleton of $\G$ correctly if and only if $\pi$ is an r-order of $\G$.
        \begin{theorem} \label{thm: Gpi}
            Suppose $\G = (\V,\E_1,\E_2)$ is a MAG and is faithful to $\PV$, and let $\Data(\V)$ be a collection of i.i.d. samples from $\PV$ with a sufficient number of samples to recover the CI relations in $\PV$.
            Then, we have the following.
            \begin{itemize}
                \item The output of Algorithm \ref{alg: Gpi} (i.e., $\Gpi$) equals the skeleton of $\G$ if and only if $\pi \in \Pi^r(\G)$.
                \item For an arbitrary order $\pi$ over set $\V$, $\Gpi$ is a supergraph of the skeleton of $\G$.
            \end{itemize}
        \end{theorem}

        Theorem \ref{thm: Gpi} implies that if $\pi \in \Pi^r(\G)$, then $\Gpi$ is the skeleton of $\G$, and if $\pi \notin \Pi^r(\G)$, then $\Gpi$ is a supergraph of the skeleton of $\G$ that contains at least one extra edge.
        Therefore, by defining the cost of an order in $\Pi(\V)$ equal to the number of edges in $\Gpi$, r-orders will be the minimizers, which implies the following.
        
        \begin{theorem}[Consistency of the score function] \label{thm: opt prob}
            Any solution of the optimization problem
            \begin{equation}\label{eq: opt problem}
                \argmin_{\pi \in \Pi(\V)} |\E^{\pi}|,
            \end{equation}
            is an r-order, i.e., a member of $\Pi^r(\G)$. Conversely every member of $\Pi^r(\G)$ is also a solution of \eqref{eq: opt problem}.
        \end{theorem}
        Next, we propose both exact and heuristic algorithms for solving the above optimization problem.

    \subsection{Algorithmic Approaches to Finding an R-order}
        In this section, we propose three algorithms for solving the optimization problem in \eqref{eq: opt problem}.
        
        \subsubsection{Hill-climbing Approach (ROL$_\text{HC}$)}
            \begin{algorithm}[t]
                \caption{Hill-climbing approach (ROL$_\text{HC}$)}
                \label{alg: hill climbing}
                \begin{algorithmic}[1]
                    \STATE \textbf{Input:} $Data(\V)$, \textit{maxSwap}, \textit{maxIter}
                    \STATE Initialize $\pi\in\Pi(\V)$ as discussed in Appendix A.1
                    \STATE $C_{\pi} \gets \textbf{ComputeCost}(\pi, \Data(\V))$
                    \FOR{ 1 to \textit{maxIter}}
                        \STATE Denote $\pi$ by $(X_1,\cdots, X_n)$
                        \STATE $\Pi^{\text{new}} \!\gets \{(X_1,\cdots X_{a-1}, X_b,X_{a+1}, \cdots, \linebreak X_{b-1}, X_a, X_{b+1}, \cdots, X_n) \vert 1 \leq b-a \leq \textit{maxSwap}\}$
                        \FOR{$\pi_{\text{new}} \in \Pi^{\text{new}}$}
                            \STATE $C_{\pi_{\text{new}}} \gets \textbf{ComputeCost}(\pi_{\text{new}}, \Data(\V))$
                            \IF{$C_{\pi_{\text{new}}} < C_{\pi}$}
                                \STATE $\pi \gets \pi_{\text{new}}$, $C_{\pi} \gets C_{\pi_{\text{new}}}$
                                \STATE \textbf{Break} go to line 5 
                            \ENDIF
                        \ENDFOR
                    \ENDFOR
                    \STATE \textbf{Return} $\pi$
                \end{algorithmic}
                \hrulefill
                \begin{algorithmic}[1]
                    \STATE \textbf{Function ComputeCost} ($\pi,\, \Data(\V)$)
                    \STATE $\Gpi=(\V,\E^{\pi}) \gets \textbf{LearnGPi}(\pi,\, \Data(\V))$
                    \STATE \textbf{Return} $|\E^{\pi}|$
                \end{algorithmic}
            \end{algorithm}
        
            In Algorithm \ref{alg: hill climbing}, we propose a hill-climbing approach, called ROL$_\text{HC}$\footnote{ROL stands for \textbf{R}-\textbf{O}rder \textbf{L}earning.} for finding an r-order.
            In general, the output of Algorithm \ref{alg: hill climbing} is a suboptimal solution to \eqref{eq: opt problem} as it takes an initial order $\pi$ and gradually modifies it to another order with less cost, but it is not guaranteed to find a minimizer of \eqref{eq: opt problem} by taking such greedy approach.
            Nevertheless, this algorithm is suitable for practice as it is scalable to large graphs, and also achieves a superior accuracy compared to the state-of-the-art methods (please refer to the experiment section).
           
            Inputs to Algorithm \ref{alg: hill climbing} are the observational data $\Data(\V)$ and two parameters \textit{maxIter} and \textit{maxSwap}.
            \textit{maxIter} denotes the maximum number of iterations before the algorithm terminates, and \textit{maxSwap} is an upper bound on the index difference of two variables that can get swapped in an iteration (line 6). 
            Initial order $\pi$ in line 2 can be any arbitrary order, but selecting it cleverly will improve the performance of the algorithm. 
            In Appendix A.1, we describe several ideas for selecting the initial order, such as initialization using the output of other approaches.
            The algorithm computes the cost of $\pi$ (denoted by $C_{\pi}$) in line 3 by calling subroutine \textit{ComputeCost} which itself calls subroutine \textit{LearnGPi} (See Algorithm \ref{alg: Gpi}).
            The remainder of the algorithm (lines 4-12) updates $\pi$ iteratively, \textit{maxIter} number of times.
            It updates the current order $\pi\!=\!(X_1,\!\cdots\!,X_n)$ as follows: first, it constructs a set of orders $\Pi^{\text{new}} \subseteq \Pi(\V)$ from $\pi$ by swapping any two variables $X_a$ and $X_b$ in $\pi$ as long as $1 \leq b-a \leq \textit{maxSwap}$.
            Next, for each $\pi_{\text{new}} \in \Pi^{\text{new}}$, it computes the cost of $\pi_{\text{new}}$ and if it has a lower cost compared to the current order, the algorithm replaces $\pi$ by that order and repeats the process.
            
            In Appendix A.2, we present a slightly modified version of Algorithm \ref{alg: hill climbing}, called Algorithm 4, which does not compute the cost of an order as in line 8 of Algorithm \ref{alg: hill climbing} but rather uses the information of $C_{\pi}$ for computing the cost of the new permutation $C_{\pi_{\text{new}}}$ (using Algorithm 3 also presented in the Appendix A.2).
            By doing so, Algorithm 4 significantly reduces the computational complexity.
            
        \subsubsection{Exact Reinforcement Learning Approach (ROL$_\text{VI}$)}
            In this section, we show that the optimization problem in \eqref{eq: opt problem} can be cast as a reinforcement learning (RL) problem.
               
            Recall the process of recovering $\Gpi$ from a given order $\pi$ in Algorithm \ref{alg: Gpi}.
            This process can be interpreted as a Markov decision process (MDP) in which the iteration index $t$ denotes time, the set of variables $\V$ represents the action space, and the state space is the set of all subsets of $\V$.
            More precisely, let $s_t$ and $a_t$ denote the state and the action of the MDP at time/iteration $t$, respectively.
            In our setting, $s_t$ is the remaining variables at time $t$, i.e., $s_t=\V_t$, and action $a_t$ is the variable that is getting removed from $\V_t$ in that iteration, i.e., $a_t=X_t$.
            Accordingly, the state transition due to action $a_t$ is $s_{t+1}=\V_t \setminus \{a_t\}$.
            The immediate reward of selecting action $a_t$ at state $s_t$ will be the negative of the instant cost, that is the number of discovered neighbors for $a_t$ by \textit{FindNeighbors} in line 6 of Algorithm \ref{alg: Gpi}, i.e., 
            \begin{align*}
                r(s_t,a_t)= |\textbf{FindNeighbors}(a_t,\, \Data(s_t))|= -|\N_{a_t}|.
            \end{align*}
            Since the form of the function $r(s_t,a_t)$ is not known, this is an RL as opposed to a classic MDP setting.
            We denote by $\pi_{\theta}$, a deterministic policy parameterized by $\theta$. 
            That is, for any state $s_t$, $a_t = \pi_{\theta}(s_t)$ is an action in $s_t$.
            Accordingly, we modify Algorithm \ref{alg: Gpi} as follows: it gets a policy $\pi_{\theta}$ instead of a permutation $\pi$ as input.
            Furthermore, it selects $X_t$ in line 5 as $X_t = \pi_{\theta}(\V_t)$.
            Given a policy $\pi_{\theta}$ and the initial state $s_1=\V$, a trajectory $\tau=(s_1,a_1,s_2,a_2,\cdots,s_{n-1},a_{n-1})$ denotes the sequence of states and actions selected by $\pi_{\theta}$.
            The cumulative reward of this trajectory, denoted by $R(\tau_{\theta})$, is the sum of the immediate rewards.
            \begin{align*}
                R(\tau_{\theta}) = \sum_{t=1}^{n-1} r(s_t,a_t) = -\sum_{t=1}^{n-1}|\N_{a_t}|.
            \end{align*}
            Hence, if we denote the output of this modified algorithm by $\G^{\theta} = (\V, \E^{\theta})$, then $R(\tau_\theta)= -|\E^\theta|$.
            In this case, any algorithm that finds the optimal policy for RL, such as Value iteration \cite{sutton2018reinforcement} or Q-learning \cite{watkins1992q} can be used to find a minimum-cost policy $\pi_\theta$.
            \begin{remark}
                According to the introduced RL setting, value-iteration can be used to find the optimal policy with the time complexity of $\mathcal{O}(n2^n)$, which is much less than $\mathcal{O}(n!)$ for naively iterating over all orders.
            \end{remark}
        \subsubsection{Approximate Reinforcement Learning Approach (ROL$_\text{PG}$)}
            Although any algorithm suited for RL is capable of finding an optimal deterministic policy for us, the complexity does not scale well as the graph size.
            Therefore, we advocate searching for a stochastic policy that increases the exploration during the training of an RL algorithm.
            As discussed earlier, we could exploit stochastic policies parameterized by neural networks to further improve scalability.
            However, this could come at the price of approximating the optimal solution instead of finding the exact one.
            In the stochastic setting, an action $a_t$ is selected according to a distribution over the remaining variables, i.e., $a_t \sim P_{\theta}(\cdot\vert s_t=\V_t)$, where $\theta$ denotes the parameters of the policy (e.g., the weights used in training of a neural network).
            In this case, the objective of the algorithm is to minimize the expected total number of edges learned by policy $P_{\theta}(\cdot\vert s_t=\V_t)$, i.e.,
            \begin{equation}\label{eq: opt problem 2}
                \argmax_{\theta} \mathbb{E}_{\tau_{\theta} \sim P_{\theta}}\big[-|\E^{\theta}|\big], 
            \end{equation}
            where the expectation is taken w.r.t. randomness of the stochastic policy.
            Many algorithms have been developed in the literature for finding stochastic policies and solving \eqref{eq: opt problem 2}.
            Some examples include Vanilla Policy Gradient (VPG) \citep{williams1992simple}, REINFORCE \citep{sutton1999policy},and Deep Q-Networks (DQN) \citep{mnih2013playing}.

\section{Second Stage: Identifying the MEC}
    In the previous section, we proposed three algorithms for finding an r-order $\pi^* \in \Pi^r(\V)$.
    Recall that our goal in this paper is to identify the MEC $[\G]$ using the available data from $\PV$.
    To this end, we can recover the skeleton of the MAGs in $[\G]$ by calling Algorithm \ref{alg: Gpi} with input $\pi^*$.
    Moreover, since \text{FindNeighbors} finds a separating set for non-neighbor variables of $X_t$ in $\G[\V_t]$, we can modify Algorithm \ref{alg: Gpi} to further return a set of separating sets for all the non-neighbor variables in MAG $\G$.
    This information suffices to identify $[\G]$ by maximally orienting the edges using the complete set of orientation rules introduced in \cite{zhang2008causal}.

\section{Experiments}
    \begin{figure*}[ht]
        \centering
        \begin{subfigure}{\textwidth}
            \centering
            \includegraphics[width=0.31\textwidth]{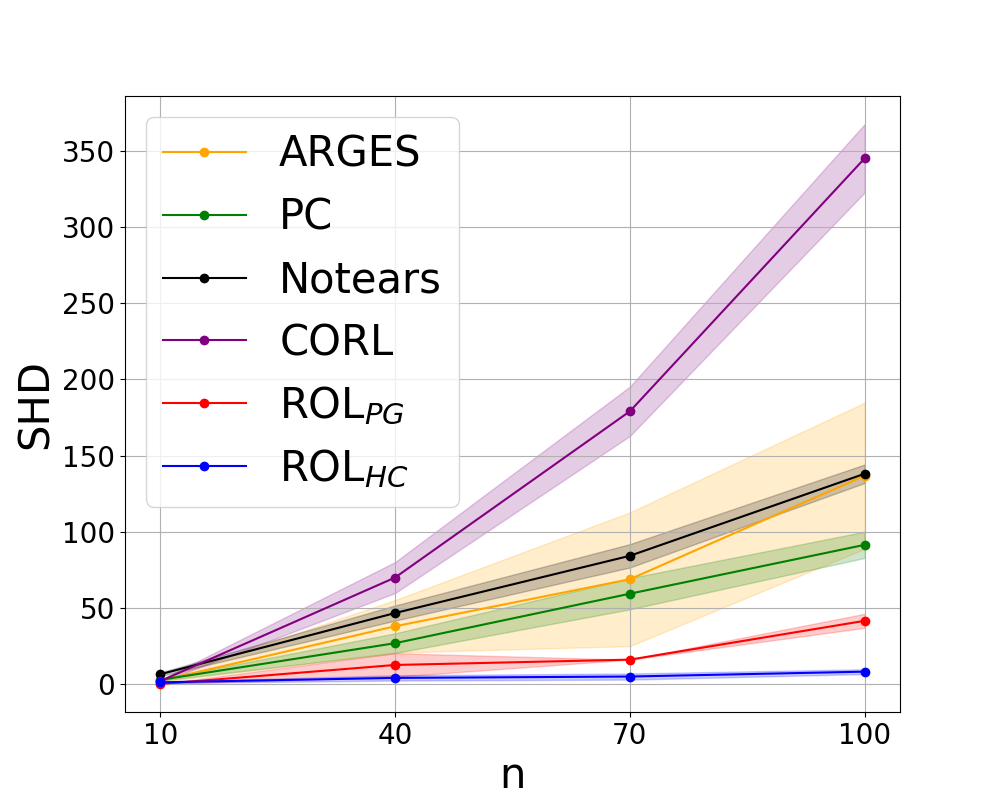}\hspace{1cm}
            \includegraphics[width=0.31\textwidth]{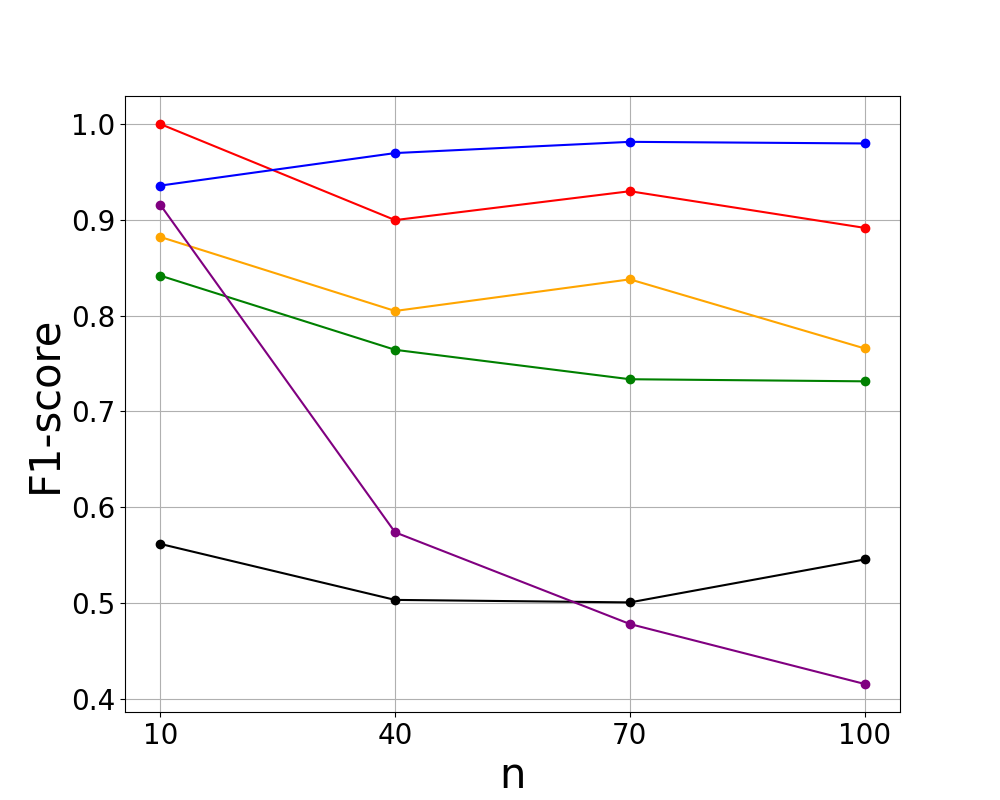}
            \caption{Erd\"os-R\`enyi $Er(n,p=n^{-0.7}),\, \text{sample size} = 50n$.}
            \label{fig: erdos}
        \end{subfigure}
        
        \begin{subfigure}{0.31\textwidth}
            \centering
            \includegraphics[width=\textwidth]{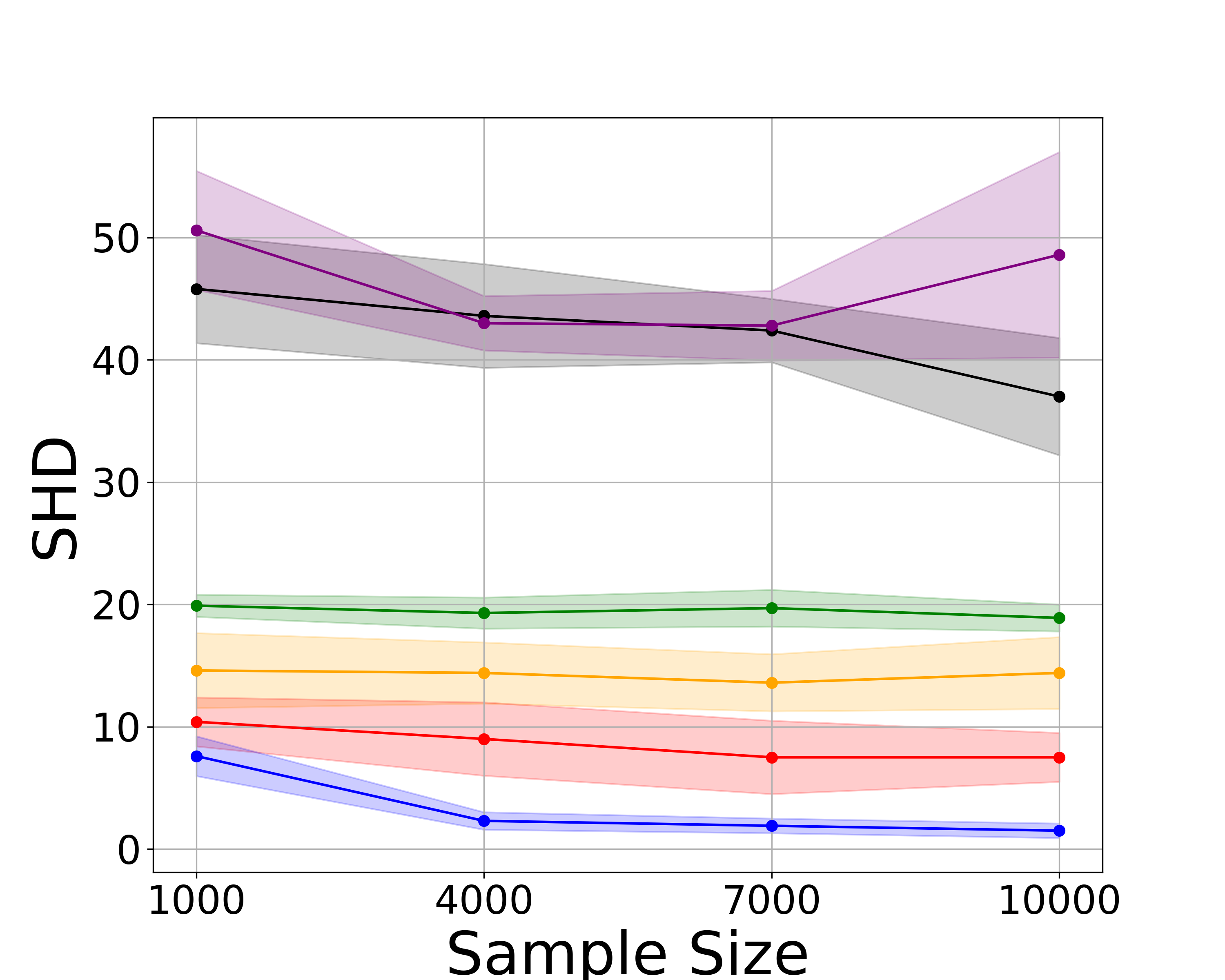}
            \caption{Alarm ($n=37, e=46$).}
            \label{fig: alarm}
        \end{subfigure}
        \begin{subfigure}{0.31\textwidth}
            \centering
            \includegraphics[width=\textwidth]{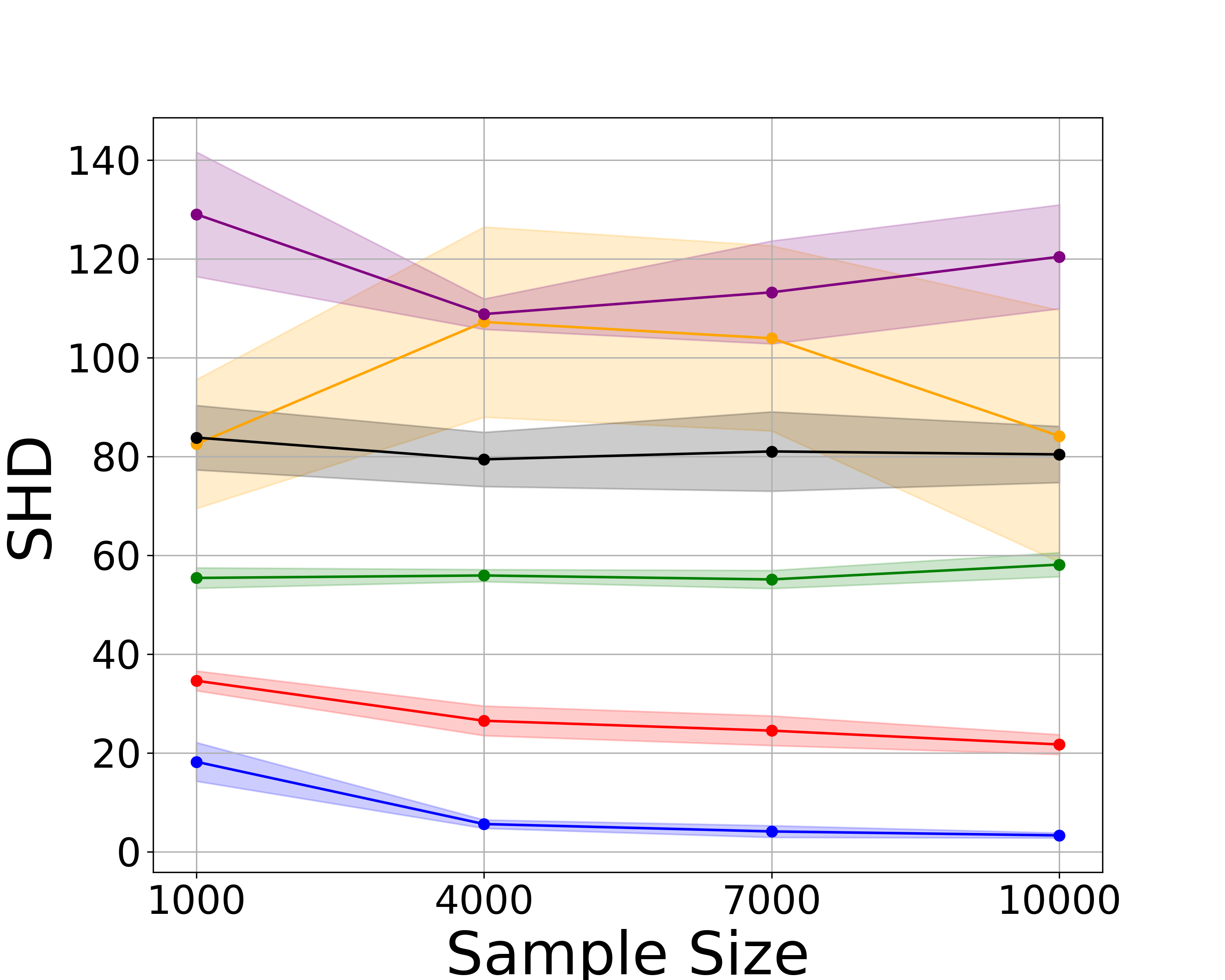}
            \caption{Barley ($n=48, e=84$).}
            \label{fig: barley}
        \end{subfigure}%
        \begin{subfigure}{0.31\textwidth}
            \centering
            \includegraphics[width=\textwidth]{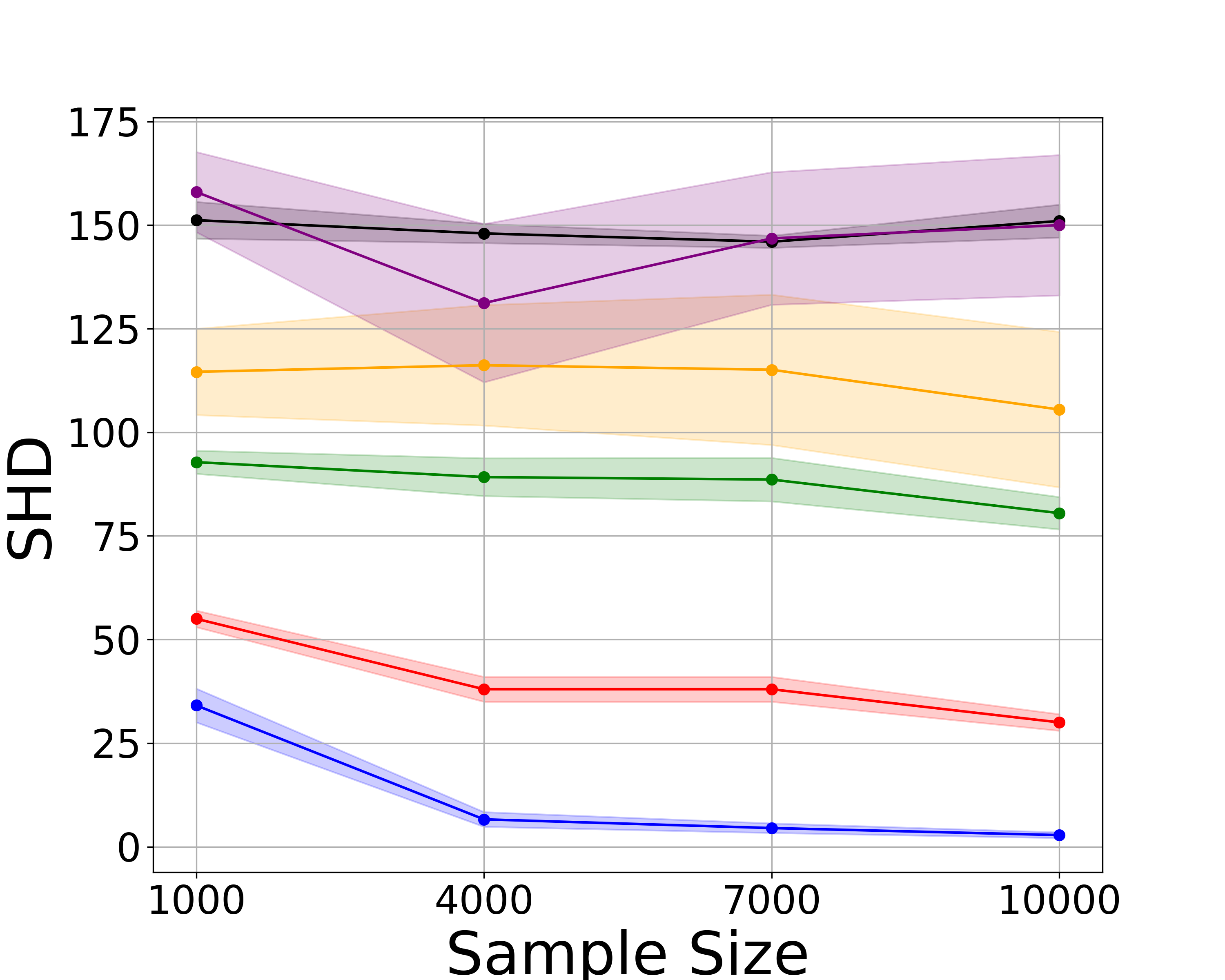}
            \caption{Hepar2 ($n=70, e=123$).}
            \label{fig: hepar2}
        \end{subfigure}
        \caption{DAG learning; $n$ and $e$ denote the number of variables and the number of edges, respectively.}
        \label{fig: DAG_learning}
    \end{figure*}

    In this section, we evaluate and compare our algorithms\footnote{Implementations of our approaches (in Python and MATLAB) are publicly available at https://github.com/Ehsan-Mokhtarian/ROL.} against two types of methods: 
    (i) those assuming causal sufficiency (DAG learning): PC \citep{spirtes2000causation}, NOTREARS \citep{zheng2018dags}, CORL \citep{wang2021ordering}, and ARGES \citep{nandy2018high}; 
    (ii) those that do not require causal sufficiency (MAG learning): RFCI \citep{colombo2012learning}, FCI+ \citep{claassen2013learning}, L-MARVEL \citep{akbari2021recursive}, MBCS* \citep{pellet2008finding}, and GSPo \citep{bernstein2020ordering}.
    
    We evaluated the aforementioned algorithms\footnote{Details pertaining to the reproducibility, hyperparameters,
and additional experiments are provided in Appendix C.} on finite sets of samples, where they were generated using a linear SEM.
    The coefficients were chosen uniformly at random from $[-1.5,-1] \cup [1,1.5]$; the exogenous noises were generated from normal distribution $\mathcal{N}(0,\sigma^2)$, where $\sigma$ was selected uniformly at random from $[0.7, 1.2]$.
    We measured the performance of the algorithms by two commonly used metrics in the literature: F1-score and Structural Hamming Distance (SHD) (the discrepancy between the number of extra and missing edges in the learned vs the ground truth graph).

    Each point on the plots is reported as the average of 10 runs with 80\% confidence interval.
    Also, each entry in the tables is reported as the average of 10 runs.
    \subsection{DAG Learning}
        We consider two types of graphs: random graphs generated from Erd\"os-R\`enyi model $Er(n,p)$ and real-world networks\footnote{http://bnlearn.com/bnrepository/}.
        To generate a DAG from $Er(n,p)$, skeleton is first sampled using the Erd\"os-R\`enyi model \citep{erdHos1960evolution} in which undirected edges are sampled independently with probability $p$. Then, the edges are oriented according to a randomly selected c-order.
        
        Figure \ref{fig: DAG_learning} shows the results for learning DAGs.
        In Figure \ref{fig: erdos}, DAGs are generated from $Er(n,p=n^{-0.7})$ and $n$ varies from 10 to 100. The size of the datasets generated for this part  is $50n$.
        Figures \ref{fig: alarm}, \ref{fig: barley}, and \ref{fig: hepar2} depict the performance of the algorithms on three real-world structures, called Alarm, Barley, and Hepar2, for a various number of samples.
        As shown in these figures, ROL$_{\text{PG}}$ and ROL$_{\text{HC}}$ outperform the state of the art in both SHD and F1-score metrics.

        \begin{table}[t]
    	    \fontsize{9}{10.5}\selectfont
    	    \centering
    	    \begin{tabular}{N M{1cm}|M{1cm}|M{1.4cm} M{0.8cm} M{0.8cm} M{1cm}}
        		&\multicolumn{2}{c|}{Structure}
     			& Earthquake
     			& Survey
     			& Asia
     			& Sachs
    			\\
    			&\multicolumn{2}{c|}{$(\#\text{nodes}, \#\text{edges})$}
     			& (5, 4)
     			& (6, 6)
     			& (8, 8)
     			& (11, 17)
    			\\
    			\hline
    			& \multirow{2}{*}{ROL$_{\text{VI}}$}
    			& F1
    			& 0.96& 1 & 0.97 & 0.97
    			\\
    			&
    			& SHD
    			& 0.4& 0 & 0.4 & 1
    			\\
    			\hline
    			& \multirow{2}{*}{ROL$_{\text{HC}}$}
    			& F1
    			& 0.96& 0.98 & 0.97 & 0.95
    			\\
    			&
    			& SHD
    			& 0.4& 0.2 & 0.4 & 1.6
    			\\
    		\end{tabular}
                \caption{Comparing ROL$_{\text{HC}}$ and ROL$_{\text{VI}}$ on small graphs.}
    	    \label{table: VI}
        \end{table}
        
        Table \ref{table: VI} illustrates the performance of ROL$_{\text{VI}}$ in comparison to ROL$_{\text{HC}}$ on four small real-world structures.
        This table shows that ROL$_{\text{VI}}$ achieves better accuracy on small graphs.
        Note that ROL$_{\text{VI}}$ unlike ROL$_{\text{HC}}$ has theoretical guarantees, but is not scalable to large graphs.
        However, ROL$_{\text{VI}}$'s performance is limited to the accuracy of CI tests, and by increasing the dataset sizes, ROL$_{\text{VI}}$ performs without any errors.

    \subsection{MAG Learning}
        \begin{table*}[!ht]
    	    \fontsize{8.5}{10}\selectfont
    	    \centering
    	    \begin{tabular}{N M{1.7cm}|M{1.5cm}| M{1.2cm} M{1.1cm} M{1.1cm} M{1.1cm} M{1.1cm} M{1.1cm} M{1.1cm}}
        		&\multicolumn{2}{c|}{Structure}
                    & Insurance
     			& Water
     			& Ecoli70
     			& Hailfinder
     			& Carpo
     			& Hepar2
     			& Arth150
    			\\
    			&\multicolumn{2}{c|}{(\#Observed, \#Unobserved)}
     			& (24, 3)
     			& (29, 3)
     			& (43, 3)
     			& (53, 3)
     			& (57, 4)
     			& (65, 5)
     			& (100, 7)
    			\\
    			\hline
    			& \multirow{2}{*}{ROL$_{\text{HC}}$}
    			& F1-score
    			& \textbf{0.89} & \textbf{0.86} & \textbf{0.93} & \textbf{0.90} & \textbf{1.00} & \textbf{0.97} & \textbf{0.93}
    			\\
    			&
    			& SHD
    			& \textbf{10.1} & \textbf{18.3} & \textbf{8.1 }& 13.9 & \textbf{0.2} & \textbf{7.6} & \textbf{19.3}
    			\\
    			\hline
    			& \multirow{2}{*}{ROL$_{\text{PG}}$}
    			& F1-score
    			& 0.86 & 0.76 & 0.90 & 0.87 & 0.97 & 0.84 & \textbf{0.93}
    			\\
    			&
    			& SHD
    			& 12.9 & 35.3 & 12.7 & 18.4 & 4.1 & 36.5 & 21.5
    			\\
    			\hline
    			& \multirow{2}{*}{RFCI}
    			& F1-score
    			& 0.74 & 0.68 & 0.84 & 0.84 & 0.86 & 0.70 & 0.86
    			\\
    			&
    			& SHD
    			& 20.5 & 34.0 & 17.7 & 19.7 & 16.8 & 52.6 & 36.3
    			\\
    			\hline
    			& \multirow{2}{*}{FCI+}
    			& F1-score
    			& 0.60 & 0.55 & 0.78 & 0.77 & 0.80 & 0.57 & 0.78
    			\\
    			&
    			& SHD
    			& 31.2 & 50.0 & 23.5 & 27.2 & 24.4 & 81.4   & 56.7
    			\\
    			\hline
    			& \multirow{2}{*}{L-MARVEL}
    			& F1-score
    			& 0.87 & 0.78 & \textbf{0.93} & \textbf{0.90} & 0.99 & 0.94 & 0.92
    			\\
    			&
    			& SHD
    			& 11.5 & 26.2 & 8.5 & \textbf{12.8} & 0.8 & 12.3 & 21.2
    			\\
    			\hline
    			& \multirow{2}{*}{MBCS*}
    			& F1-score
    			& 0.77 & 0.62 & 0.90 & 0.83 & 0.99 & 0.92 & 0.87
    			\\
    			&
    			& SHD
    			& 17.8 & 38.7 & 12.0 & 20.1 & 1.1 & 17.0 & 34.2
    			\\
    			\hline
    			& \multirow{2}{*}{GSPo}
    			& F1-score
    			& 0.75 & 0.60 & 0.66 & 0.58 & 0.84 & 0.58 & 0.45
    			\\
    			&
    			& SHD
    			& 32.3 & 89.1 & 67.6 & 101.4 & 31.2 & 170.6 & 358.5
    			\\
    			\end{tabular}
                \caption{MAG learning; performance of various algorithms on seven real-world structures, when sample size is $50n$.}
    	    \label{table: MAG learning}
        \end{table*}

        We selected seven real-world DAGs for this part.
        For each structure, we randomly removed $5\%$ to $10\%$ of the variables and constructed a MAG over the set of observed variables (those not eliminated) using the latent projection approach of \citep{verma1991equivalence}. 
        Finally, we generated a finite set of samples over all the variables and fed the data pertaining to the observed variables as the input to all the algorithms.
        The goal of all algorithms is to learn the MEC of the corresponding MAG from the samples they have.

        Table \ref{table: MAG learning} presents the results.
        As demonstrated by the bold entries in the table, ROL$_{\text{HC}}$ achieves the best F1-score and SHD in almost all the cases.
        \begin{remark}
            Recall that prior to this work, GSPo was the only ordering-based method in the literature that does not require causal sufficiency.
            However, the table shows that it has the worst performance among the algorithms and is not scalable to large graphs.
            For instance, it has a poor performance on Arth150, which is a graph with 100 variables.
        \end{remark}

\section{Conclusion, Limitations, Future work}
    We advocated for a novel type of order, called an r-order, and argued that r-orders are advantageous over the previously used orders in the literature.
    Accordingly, we proposed three algorithms for causal structure learning in the presence of unobserved variables:
    ROL$_{\text{HC}}$, a Hill-climbing-based heuristic algorithm that is scalable to large graphs;
    ROL$_{\text{VI}}$, an exact RL-based algorithm that has theoretical guarantees but is not scalable to large graphs;
    ROL$_{\text{PG}}$, an approximate RL-based algorithm that exploits stochastic policy gradient.
    We showed in our experiments that ROL$_{\text{VI}}$ on small graphs and ROL$_{\text{HC}}$ on larger graphs outperform the state-of-the-art algorithms.
    Although ROL$_{\text{PG}}$ is scalable to large graphs and outperforms the existing methods, ROL$_{\text{HC}}$ performs slightly better, mainly due to better initialization.
    The weights of the neural networks in ROL$_{\text{PG}}$ are selected randomly, while we proposed clever methods for the initialization step in ROL$_{\text{HC}}$.
    Nevertheless, an important future work is to improve the policy gradient approaches.

\bibliography{bibliography}

\clearpage
\include{Appendix}

\end{document}

%% file: Appendix.tex
\appendix
\onecolumn
\begin{center}
    \bfseries\Large Appendix
\end{center}

\section{A \quad Implementation of the Hill-climbing Approach}
    In this section, we discuss some details regarding implementing our hill-climbing approach.
    In Section A.1, we introduce three ways to initialize $\pi$ in line 2 of Algorithm \ref{alg: hill climbing}.
    In Section A.2, we present a slightly modified version of Algorithm \ref{alg: hill climbing} which does not compute the cost of an order as in line 8 of Algorithm \ref{alg: hill climbing} but rather uses the information of $C_{\pi}$ for computing the cost of the new permutation $C_{\pi_{\text{new}}}$.

    \subsection{A.1 \quad Initialization in Line 2 of Algorithm \ref{alg: hill climbing}}
        \paragraph{Random Permutation:}
            This is a naive approach in which we select the initial order randomly from $\Pi(\V)$.
        
        \paragraph{Sort by the Size Markov Boundary:}
            The Markov boundary of a variable is defined in the following.
           \begin{definition}[Markov boundary]
                Markov boundary of $X\in \V$, denoted by $\Mb{X}{\V}$, is a minimal set of variables $\Z \subseteq \V \setminus \{X\}$, such that $X$ is independent of the rest of the variables of $\V$ conditioned on $\Z$.
            \end{definition}
            There exist several algorithms in the literature for discovering the Markov boundaries \citep{margaritis1999bayesian, pellet2008using}. 
            For instance, TC \cite{pellet2008using} algorithm proves that $X$ and $Y$ are in each other's Markov boundary if and only if
            \begin{equation*}
                \nsep{X}{Y}{\V \setminus \{X,Y\}}{\PV}.
            \end{equation*}
            The following lemma shows that removable variables have a small Markov boundary size.
            \begin{lemma}[Lemma 3 in \citep{akbari2021recursive}]
                If $X$ is removable in $\G$, then for every $Y\in \Mb{X}{\V}$, we have 
                \begin{equation}
                    \Mb{X}{\V} \subseteq \Mb{Y}{\V}.
                \end{equation}
            \end{lemma}
            Motivated by this lemma, we initialize $\pi$ by the sorting of $\V$ based on the Markov boundary size in ascending order.
            That is, $\pi := (X_1,\cdots, X_n)$, where $\Mb{X_1}{\V} \leq \cdots \leq \Mb{X_n}{\V}$.

        \paragraph{A recursive Approaches:}
            As mentioned in the main text, some recursive approaches have been proposed in the literature of structure learning \citep{mokhtarian2021recursive, akbari2021recursive, mokhtarian2021learning}.
            These methods identify a removable variable in each iteration by performing a set of CI tests.
            Herein, we use one of those algorithms and store the order of removed variables.
            In practice, the CI tests are noisy.
            Accordingly, their removability test may not be accurate.
            However, we can use their found order to initialize $\pi$.
            By doing so, and as illustrated in our experiments, the accuracy of our proposed method is significantly improved.
        
    \subsection{A.2 \quad Computing the Cost of a New Permutation}
        Herein, we propose Algorithm \ref{alg: compute cost}, an alternative way for computing the cost of an order $\pi=(X_1,\cdots, X_n)$ instead of using Algorithm \ref{alg: Gpi} and counting the number of edges in $\Gpi$.
        
        \begin{algorithm}[ht]
            \caption{Compute cost of an order}
            \label{alg: compute cost}
            \begin{algorithmic}[1]
                \STATE \textbf{Function ComputeCost} ($\pi,\,a,\,b,\, \Data(\V)$)
                \STATE $\Vrem \gets \V$
                \STATE $C_{ab} \gets (0,0,\cdots,0) \in \mathbb{R}^{n}$
                \FOR{$t=a$ to $b$}
                    \STATE $X \gets \pi(t)$
                    \STATE $\N_X \gets \textbf{FindNeighbors}(X,\, \Data(\Vrem))$
                    \STATE $\C_{ab}(t) \gets |\N_X|$
                    \STATE $\Vrem \gets \Vrem \setminus \{X\}$
                \ENDFOR
                \STATE \textbf{Return} $C_{ab}$
            \end{algorithmic}
        \end{algorithm}
        
        This algorithm additionally gets two numbers $1\leq a<b \leq n$ as input and returns a vector $C_{ab} \in \mathbb{R}^{n}$.
        For $a\leq i \leq b$, the $i$-th entry of $C_{ab}$ is equal to $|N_{X_i}|$ which is the number of neighbors of $X_i$ in the remaining graph.
        Hence, to learn the total cost of $\pi$, we can call this function with $a=1$ and $b=n$ and then compute the sum of the entries of the output $C_{1n}$.
        Accordingly, we modify Algorithm \ref{alg: hill climbing} and propose Algorithm \ref{alg: modified hill climbing}.
        
        \begin{algorithm}[ht]
            \caption{Modified hill climbing approach}
            \label{alg: modified hill climbing}
            \begin{algorithmic}[1]
                \STATE \textbf{Input:} $Data(\V)$, \textit{maxSwap}, \textit{maxIter}
                \STATE Initialize $\pi\in\Pi(\V)$ as discussed in Appendix A.1
                \STATE $C_{1n} \gets \textbf{ComputeCost}(\pi,1,n, \Data(\V))$
                \FOR{ 1 to \textit{maxIter}}
                    \STATE Denote $\pi$ by $(X_1,\cdots, X_n)$
                    \FOR{$1\leq a < b \leq n$ such that $b-a< \text{maxSwap}$}
                        \STATE $\pi_{\text{new}} \gets$ Swap $X_a$ and $X_b$ in $\pi$
                        \STATE $C^{\text{new}}_{ab} \gets \textbf{ComputeCost}(\pi_{\text{new}},a,b, \Data(\V))$
                        \IF{$\sum_{i=a}^b C^{\text{new}}_{ab}(i) < \sum_{i=a}^b C_{1n}(i)$}
                            \STATE $\pi \gets \pi_{\text{new}} $
                            \STATE $C_{1n}(a:b) \gets C^{\text{new}}_{ab}(a:b)$
                            \STATE \textbf{Break} the for loop in line 5
                        \ENDIF
                    \ENDFOR
                \ENDFOR
                \STATE \textbf{Return} $\pi$
            \end{algorithmic}
        \end{algorithm}

        In this algorithm, $C_{1n}$ is a vector that stores the cost of $\pi$, computed in line 3.
        Similar to Algorithm \ref{alg: hill climbing}, a new permutation $\pi_{\text{new}}$ is constructed by swapping two entries $X_a$ and $X_b$ in $\pi$.
        The main change is in line 8, where the algorithm calls function \textit{ComputeCost} with $a$ and $b$ to compute the cost of the new policy.
        The reason is that for $i<a$ and $i>b$, the $i$-th entry of the cost of $\pi$ and $\pi_{\text{new}}$ are the same. This is because the set of remaining variables is the same.
        Hence, to compare the cost of $\pi$ with the cost of $\pi_{\text{new}}$, it suffices to compare them for entries between $a$ and $b$.
        Accordingly, Algorithm \ref{alg: modified hill climbing} checks the condition in line 9, and if the cost of the new policy is better, then the algorithm updates $\pi$ and its corresponding cost.
        Note that it suffices to update the entries between $a$ to $b$ of $C_{1n}$.

\section{B \quad Technical proofs}
    \begin{customprp}{\ref{prp: not learnable}}
        Let $\G$ denotes a DAG with MEC $[\G]^{\text{DAG}} = \{\G_1,...,\G_k\}$. 
        For any two distinct DAGs $\G_i$ and $\G_j$ in $[\G]^{\text{DAG}}$, we have $\Pi^c(\G_i) \cap \Pi^c(\G_j) = \varnothing$.
    \end{customprp}
    \begin{proof}
        Let $\G_i = (\V,\E_i)$ and $\G_j = (\V,\E_j)$ denote two distinct DAGs in $[\G]^{\text{DAG}}$.
        It is well-known that two Markov equivalent DAGs have the same skeleton \citep{verma1991equivalence}.
        Hence, $\G_i$ and $\G_j$ have the same skeleton.
        Moreover, as $\G_i$ and $\G_j$ are two distinct DAGs, there exist two variables $X$ and $Y$ in $\V$ such that $(X,Y) \in \E_i$ and $(Y,X) \in \E_j$.
        Recall that in c-orders, parents of a variable appear after that variable in the order.
        This implies that in any c-order of $\G_i$, $X$ appears after $Y$, but in any c-order of $\G_j$, $X$ appears before $Y$.
        Therefore, the set of c-orders of $\G_i$ and $\G_j$ do not share any common order, i.e., $\Pi^c(\G_i) \cap \Pi^c(\G_j) = \varnothing$.
    \end{proof}

    \begin{customprp}{\ref{prp: r-order is learnable}}
        If $\G_1$ and $\G_2$ are two Markov equivalent MAGs, then $\Pi^r(\G_1) = \Pi^r(\G_2)$.
    \end{customprp}
    \begin{proof}
        To prove this proposition, we first provide a corollary of a result in \citep{akbari2021recursive}.
        \begin{corollary}[It follows from Theorem 2 in \citep{akbari2021recursive}] \label{cor: removable in MEC}
            For two Markov equivalent MAGs $\G^1$ and $\G^2$ over $\V$ and an arbitrary variable $X\in \V$, $X$ is removable in $\G^1$ if and only if  $X$ is removable in $\G^2$.
        \end{corollary}
    
        Suppose $\pi = (X_1,\cdots, X_n) \in \Pi^r(\G_1)$.
        It suffices to show that $\pi \in \Pi^r(\G_2)$, i.e., $X_i$ is removable in $\G_2[\{X_i, \cdots, X_n\}]$ for each $1 \leq i \leq n$.
        
        By induction on $i$, we show that $\G_1[\{X_i,\cdots,X_n\}]$ and $\G_2[\{X_i,\cdots,X_n\}]$ are Markov equivalent.
        This claim holds for $i=1$ due to the assumption of the proposition.
        Suppose it holds for an $i<n$ and we want to show that $\G_1[X_{i+1},\cdots,X_n]$ and $\G_2[\{X_{i+1},\cdots,X_n\}]$ are Markov equivalent.
        As $\pi \in \Pi^r(\G_1)$, $X_i$ is removable in $\G_1[\{X_i, \cdots, X_n\}]$.
        Furthermore, Corollary \ref{cor: removable in MEC} implies that $X_i$ is also removable in $\G_2[\{X_i, \cdots, X_n\}]$.
        Therefore, for any variables $Y,W\in \{X_{i+1},\cdots, X_n\}$ and $\Z \subseteq \{X_{i+1},\cdots, X_n\} \setminus \{Y,W\}$, we have
        \begin{equation} \label{eq: prp 2, eq 1}
            \sep{Y}{W}{\Z}{\G_1[X_i,\cdots,X_n]}
            \iff
            \sep{Y}{W}{\Z}{\G_1[X_{i+1},\cdots,X_n]},
        \end{equation}
        and 
        \begin{equation} \label{eq: prp 2, eq 2}
            \sep{Y}{W}{\Z}{\G_2[X_i,\cdots,X_n]}
            \iff
            \sep{Y}{W}{\Z}{\G_2[X_{i+1},\cdots,X_n]}.
        \end{equation}
        Because $\G_1[\{X_i,\cdots,X_n\}]$ and $\G_2[\{X_i,\cdots,X_n\}]$ are Markov equivalent, they impose the same m-separation relations over $\{X_i,\cdots,X_n\}$.
        Thus, Equations \eqref{eq: prp 2, eq 1} and \eqref{eq: prp 2, eq 2} imply that $\G_1[\{X_{i+1},\cdots,X_n\}]$ and $\G_2[\{X_{i+1},\cdots,X_n\}]$ impose the same m-separation relations over $\{X_{i+1},\cdots,X_n\}$, i.e., they are Markov equivalent which proves our claim.

        So far, we have shown that $\G_1[\{X_i,\cdots,X_n\}]$ and $\G_2[\{X_i,\cdots,X_n\}]$ are Markov equivalent for each $1 \leq i \leq n$.
        In this case, for each $1 \leq i \leq n$, Corollary \ref{cor: removable in MEC} implies that $X_i$ is removable in $\G_2[\{X_i, \cdots, X_n\}]$, because it is removable in $\G_1[\{X_i, \cdots, X_n\}]$.
        This completes the proof.
    \end{proof}

    \begin{customprp}{\ref{prp: c-o subset of r-o}}
        For any DAG $\G$, we have $\Pi^c(\G) \subseteq \Pi^r(\G)$.
    \end{customprp}
    \begin{proof}
        Let $\pi=(X_1,\cdots,X_n)$ be an order in $\Pi^c(\G)$.
        It suffices to show that $\pi \in \Pi^r(\G)$ or equivalently, $X_i$ is removable in DAG $\G[\{X_i,\cdots, X_n\}]$ for each $1 \leq i \leq n$.
        Let $1 \leq i \leq n$ be arbitrary.
        As $\pi \in \Pi^c(\G)$, there is no directed edge from $X_i$ to the variables in $\{X_{i+1},\cdots, X_n\}$.
        Hence, $X_i$ has not child in $\G[\{X_i,\cdots, X_n\}]$.
        In this case, Remark 6 in \citep{mokhtarian2021recursive} proves that $X_i$ is removable in DAG $\G[\{X_i,\cdots, X_n\}]$.
    \end{proof}

    \begin{customthm}{\ref{thm: Gpi}}
        Suppose $\G = (\V,\E_1,\E_2)$ is a MAG and is faithful to $\PV$, and let $\Data(\V)$ be a collection of i.i.d. samples from $\PV$ with a sufficient number of samples to recover the CI relations in $\PV$.
        Then, we have the following.
        \begin{enumerate}
            \item The output of Algorithm \ref{alg: Gpi} (i.e., $\Gpi$) equals the skeleton of $\G$ if and only if $\pi \in \Pi^r(\G)$.
            \item For an arbitrary order $\pi$ over set $\V$, $\Gpi$ is a supergraph of the skeleton of $\G$.
        \end{enumerate}
    \end{customthm}
    \begin{proof}
        We prove the two parts of the theorem separately.
        \begin{enumerate}
            \item 
            Suppose $\G^1$ is a MAG over a set $\V^1$ that is faithful to a distribution $P_{\V^1}$ and let $\V^2$ be an arbitrary subset of $\V^1$.
            The latent projection of $\G^1$ over $\V^2$ is a MAG over $\V^2$ that is faithful to $P_{\V^2}$.
            It is well-known that $\G^2$ is a supergraph of $\G^1$ where some extra edges are added to $\G^1$  according to some new unblocked paths, known as \textit{inducing paths} \citep{verma1991equivalence}.
            Equipped with this definition, \citep{akbari2021recursive} provides the following result.
            
            \begin{lemma}[Proposition 1 in \citep{akbari2021recursive}] \label{lem: proof of thm}
                Suppose $\G^1$ is a MAG over a set $\V^1$ that is faithful to $P_{\V^1}$.
                Let $X \in \V^1$ be arbitrary and $\G^2$ be the latent projection of $\G^1$ over $\V^2$, where $\V^2 = \V^1 \setminus \{X\}$.
                In this case, MAG $\G^1[\V^2]$ is equal to $\G^2$ if and only if $X$ is removable in $\G^1$.
            \end{lemma}
            Using this lemma, we complete the proof in the following.
            
            \textit{Sufficient part:}
            Suppose $\pi=(X_1,\cdots,X_n) \in \Pi^r(\G)$, i.e., $X_t$ is removable in $\G[\{X_t,\cdots,X_n\}]$ for each $1 \leq t \leq n$.
            Recall that $\V_t = \{X_t,\cdots,X_n\}$ in Algorithm \ref{alg: Gpi}.
            Lemma \ref{lem: proof of thm} implies that at each iteration $t$, $\G[\V_t]$ is faithful to $P_{\V_t}$.
            In this case, function \textit{FindNeighbors} correctly learns the neighbors of $X_t$ among the remaining variables $\V_t$, i.e., $N_{X_t}$ is the set of neighbors of $X_t$ in $\G[\V_t]$.
            Hence, $\G^{\pi}$ equals the skeleton of $\G$ at the end of the algorithm.
            
            \textit{Necessary part:} 
            Suppose $\pi=(X_1,\cdots,X_n) \notin \Pi^r(\G)$, i.e., there exists $1 \leq i \leq n$ such that $X_i$ is not removable in $\G[\V_i]$.
            Recall that $\V_i = \{X_i,\cdots, X_n\}$ and $\V_{i+1} = \V_i \setminus \{X_i\}$.
            Let $\G^1$ denotes $\G[\V_i]$ and $\G^2$ denotes the latent projection of $\G^1$ on $\V_{i+1}$.
            In this case, Lemma \ref{lem: proof of thm} implies that $\G^2$ has some extra edges compared to $\G^1[\V_{i+1}] =  \G[\V_{i+1}]$.
            Let $(Y,W)$ be an edge in $\G^2$ that is not in $\G[\V_{i+1}]$.
            In this case, $Y$ and $W$ are not m-separable using the subsets of $\V_{i+1}$.
            Hence, $\{Y, W\}$ will be added to the learned graph $\G^{\pi}$.
            Note that this edge does not exist in $\G$, which shows that the learned graph is not equal to the skeleton of $\G$.
            \item 
            Let $\pi = \{X_1,\cdots, X_n\}$ and $(X_a,X_b)$ be an edge in $\G$.
            It suffices to show that $\{X_a, X_b\}$ appears in $\Gpi$.
            When $t=\min(a,b)$ in Algorithm \ref{alg: Gpi}, both $X_a$ and $X_b$ exists in $\V_t$.
            Note that $X_t \in \{X_a,X_b\}$.
            In this case, $X_a$ and $X_b$ are not m-separable as they are neighbors.
            Hence, function \textit{FindNeighbors} identify them as neighbors and therefore, undirected edge $\{X_a,X_b\}$ will be added to $\Gpi$.
        \end{enumerate}
    \end{proof}

\section{C \quad Complementary of the Experiment Section}
    In this section, we provide additional experiments, discuss details pertaining to the reproducibility and hyper-parameters, formally define the evaluation metrics, illustrate the runtime and discuss the time-complexity of our methods, and present more details about the real-world structures used in our experiments.
    
    \subsection{C.1 \quad Additional Experiments}
        \subsubsection{Additional Experiments on Random Graphs:}
\begin{figure}[ht]
                \centering
                \begin{subfigure}{\textwidth}
                    \centering
                    \includegraphics[width=0.4\textwidth]{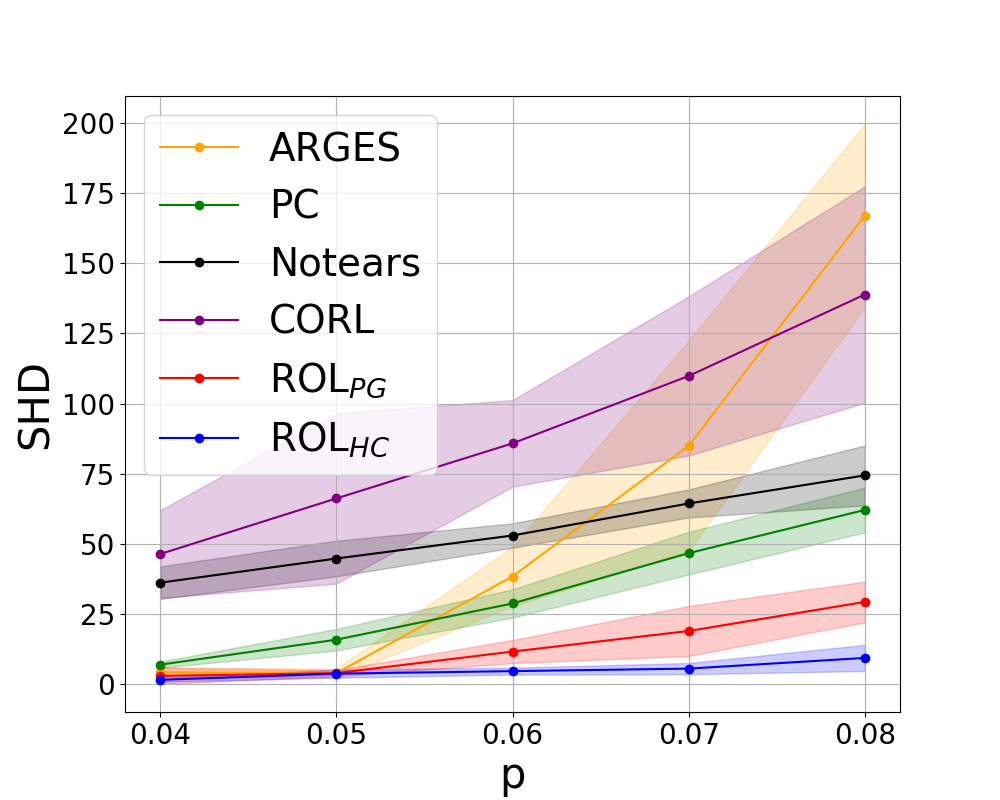}\hspace{1cm}
                    \includegraphics[width=0.4\textwidth]{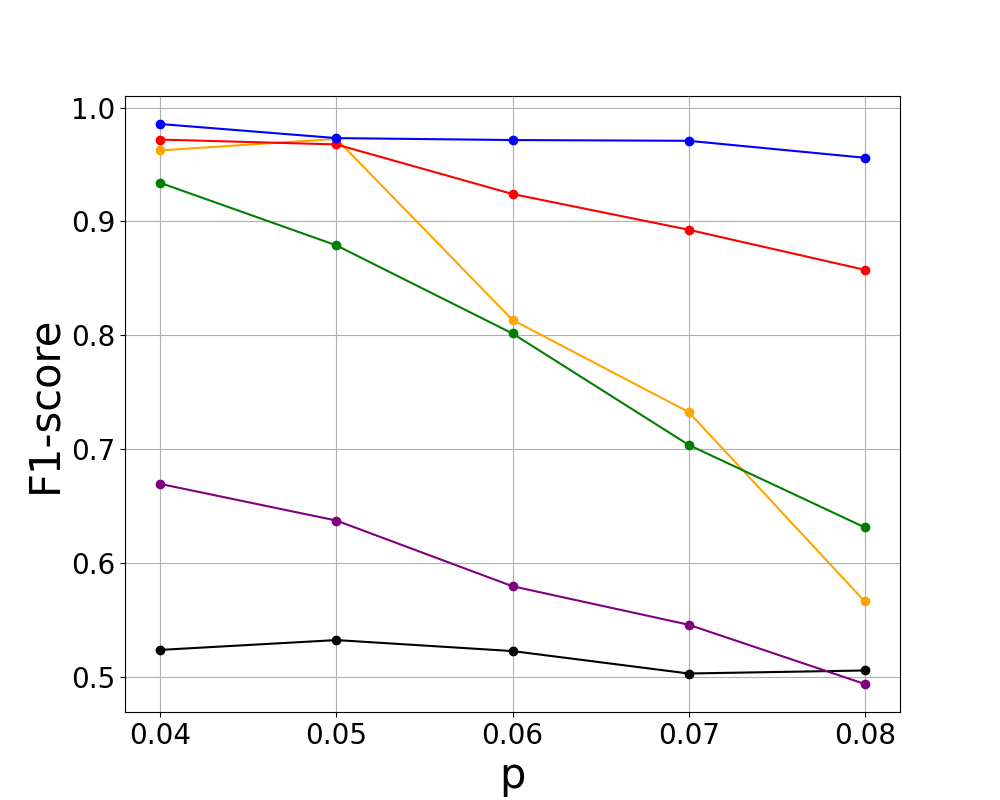}
                \end{subfigure}
                \caption{Erd\"os-R\`enyi $Er(n=50, p),\, \text{sample size} = 2500$.}
                \label{fig: DAG apd}
            \end{figure}

            To illustrate the effect of the graph density on the performance of various learning algorithms, in this section, we evaluated several algorithms on randomly generated graphs with a fixed number of vertices and different densities.
            Figure \ref{fig: DAG apd} illustrates the performance of various learning algorithms on random graphs generated from $Er(50,p)$ model when $n=50$ and $p$ varies from $0.04$ to $0.08$.
            Similar to Figure \ref{fig: 1a} in the main text, ROL$_{\text{PG}}$ and ROL$_{\text{HC}}$ outperform the state of the art in both SHD and F1-score metrics.

        \subsubsection{ROC Curves}
\begin{figure}[ht]
                \centering
                \begin{subfigure}{\textwidth}
                    \centering
                    \includegraphics[width=0.7\textwidth]{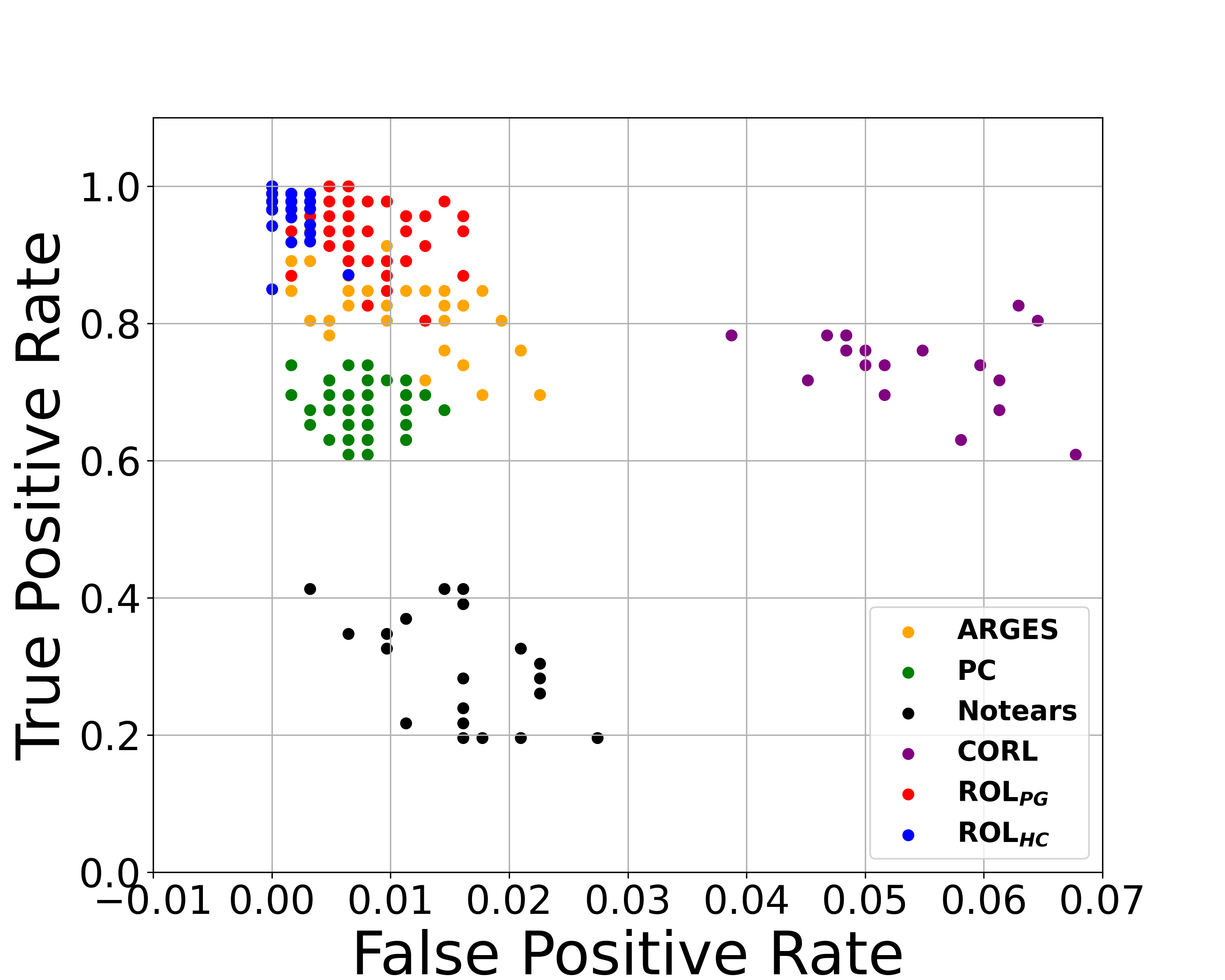}
                \end{subfigure}
                \caption{ROC plot of various algorithms for learning graph Alarm.}
                \label{fig: DAG roc}
            \end{figure}

            we present the ROC of different methods in Figure \ref{fig: DAG roc} for learning the graph of dataset Alarm.
            This shows that our algorithms have both higher true positive (TP) and lower false positive (FP) rates compared to the others.

    \subsection{C.2 \quad Reproducibility and Hyper-parameters}
        For our experiments, we utilized a Linux server with Intel Xeon CPU E5-2680 v3 (24 cores) operating at 2.50GHz with 256 GB DDR4 of memory and 2x Nvidia Titan X Maxwell GPUs. 
    
        Next, we introduce all the details, including hyper-parameters that were used in the implementation of our proposed methods.

        \subsubsection{ROL$_{\text{HC}}$:}
            In our implementation, we used the modified version of our algorithm introduced in Appendix A.2, i.e., Algorithms \ref{alg: compute cost} and \ref{alg: modified hill climbing}.
            We called the algorithms with parameters $\textit{maxIter} = 20$ and $\textit{maxSwap} = 10$.
            To initialize $\pi$ in line 2, we used the recursive approach introduced in Appendix A.1 by the method in \citep{mokhtarian2021learning}.
            As discussed in the main text, there are several algorithms in the literature to implement \textit{FindNeighbors} function.
            Herein, we used the algorithm in \citep{mokhtarian2021learning} which is reasonably fast for large graphs.
            To perform CI tests from the available dataset, we used Fisher Z-transformation with a significance level of $\alpha= 0.01$.
           Please note that we repeated almost all the experiments with different $\alpha$, e.g., $\alpha = 0.05$, and the results remained very similar to the results for $\alpha = 0.01$.
            Thus, we did not report them separately.
        
        \subsubsection{ROL$_{\text{PG}}$:}
            For our RL approach, we used Garage library \citep{garage} as it is easy to maintain and parallelize. 
            To this end, we implemented an environment class compatible with Garage library and Softmax Policy.
            We used an MLP with two hidden layers, each with 64 neurons, and the $\tanh$ activation function for the hidden layers to model the policy function.
            Afterward, we employed a modified version of the Vanilla Policy Gradient algorithm that allows us to fit a linear feature baseline in order to achieve better performance by reducing the variance \cite{sutton1999policy}.
        
    \subsection{C.3 \quad Evaluation Metrics}
        As mentioned in the experiment section, we measured the performance of the algorithms by two commonly used metrics in the literature: F1-score and Structural Hamming Distance (SHD).
        Herein, we formally define these measures.
        
        Let $\G_1$ and $\G_2$ denote the true graph's skeleton and the learned graph's skeleton, respectively.
        We first define a few notations.
        True-positive (TP) is the number of edges that appear in both $\G_1$ and $\G_2$.
        False-positive (FP) is the number of edges that appear in $\G_2$ but do not exist in $\G_1$.
        False-negative (FN) is the number of edges in $\G_1$ that the algorithm failed to learn in $\G_2$.
        In this case, SHD is defined as follows.
        \begin{equation*}
            \text{SHD = FP + FN}.
        \end{equation*}
        To define F1-score, we first define \textit{precision} and \textit{recall} in the following.
        \begin{equation*}
            \text{Precision = } \frac{\text{TP}}{\text{TP + FP}}.
        \end{equation*}
        \begin{equation*}
            \text{Recall = } \frac{\text{TP}}{\text{TP + FN}}.
        \end{equation*}
        Finally, F1-score is defined as follows.
        \begin{equation*}
            \text{F1-score = } 2 \times \frac{\text{Precision }\times \text{ Recall} }{\text{Precision }+ \text{ Recall}}.
        \end{equation*}
        Note that F1-score is a number between 0 to 1, while larger numbers indicate better accuracy.

    \subsection{C.4 \quad Time Complexity and Runtime}
\begin{figure}[ht]
            \centering
            \begin{subfigure}{\textwidth}
                \centering
                \includegraphics[width=0.35\textwidth]{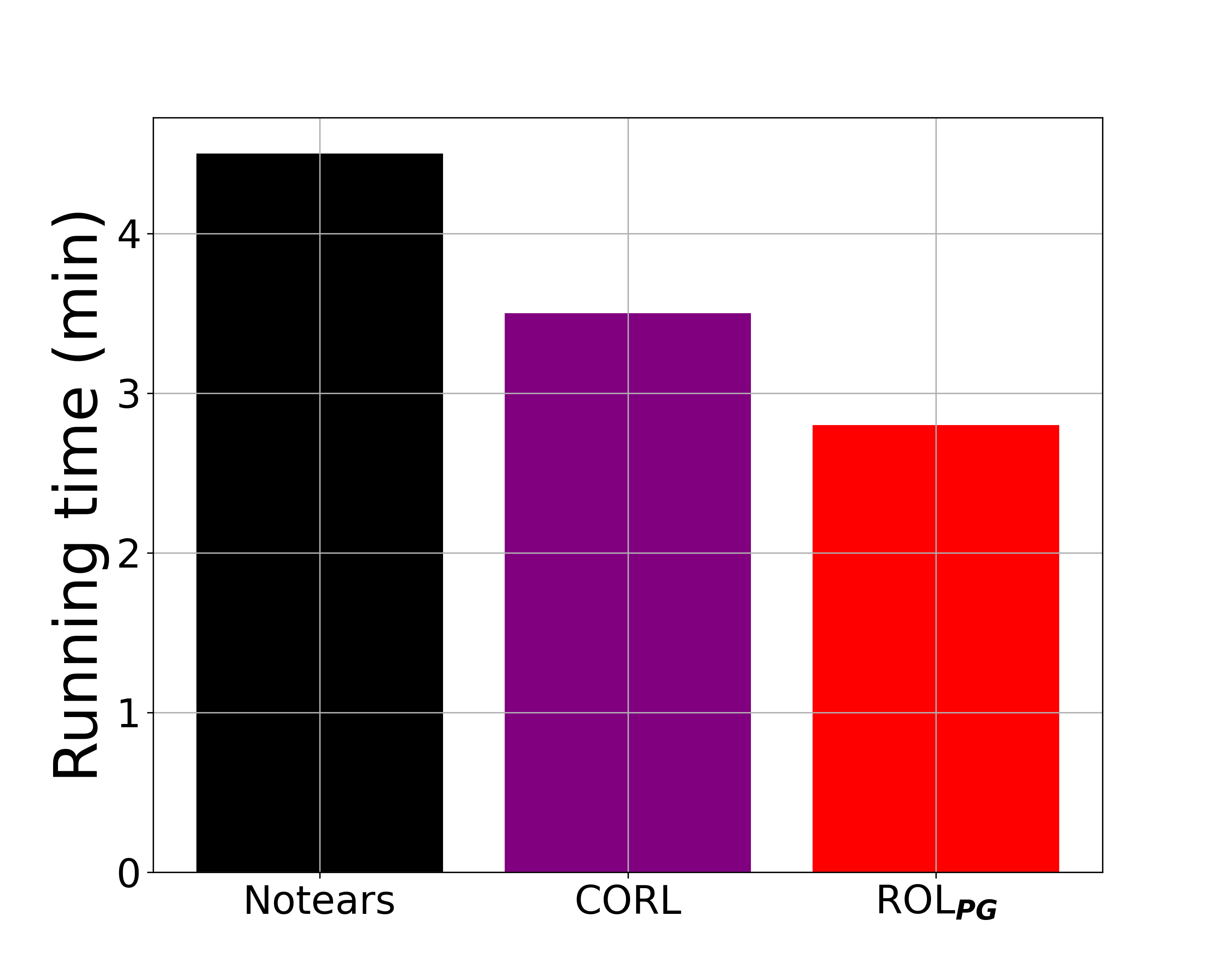}\hspace{1cm}
                \includegraphics[width=0.35\textwidth]{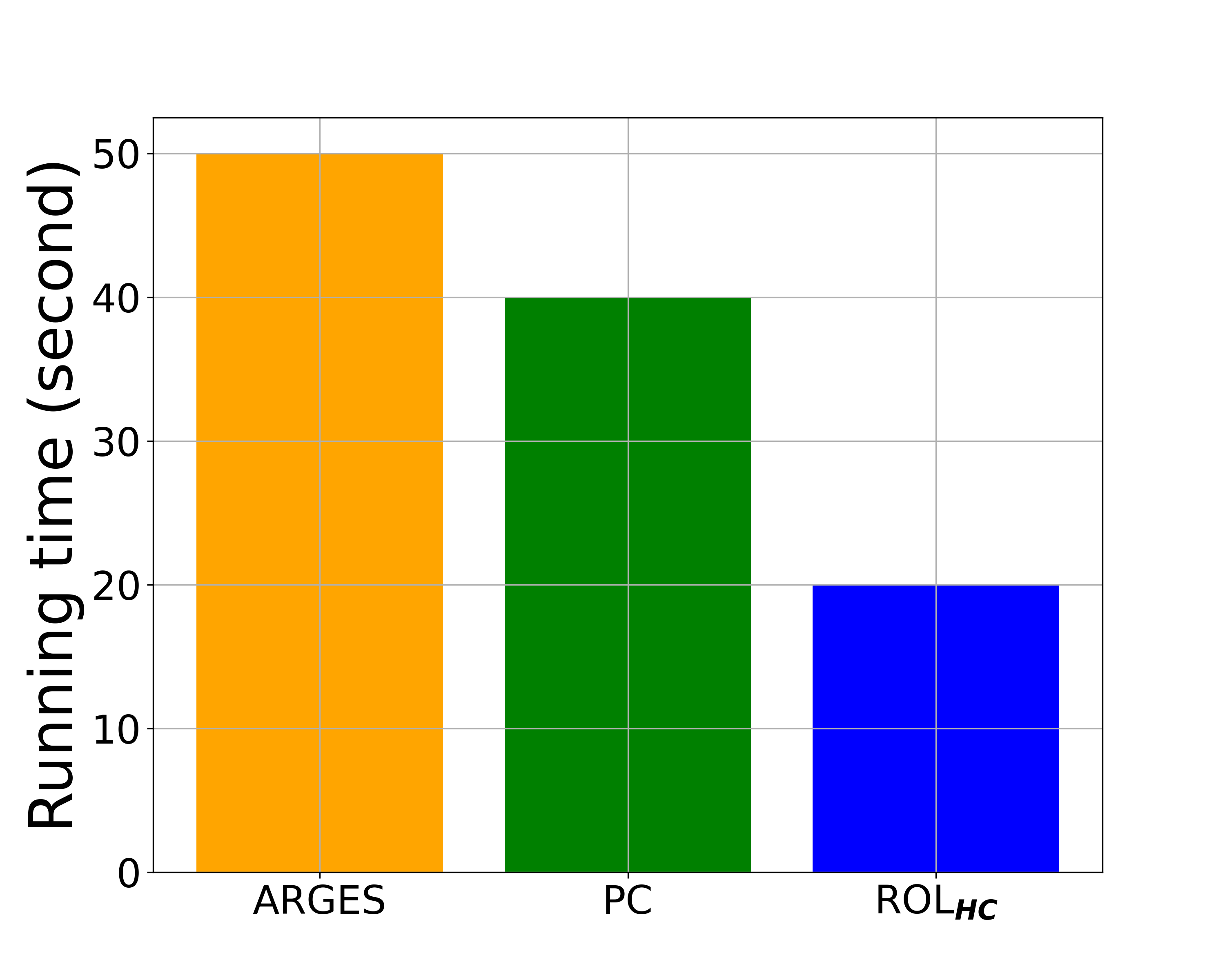}
            \end{subfigure}
            \caption{Runtime of various algorithms for learning graph Barley. }
            \label{fig: DAG time_compare}
        \end{figure}

        We illustrate the runtime of various algorithms in Figure \ref{fig: DAG time_compare} for learning the graph of the dataset Barley.
        This figure shows that our algorithms are relatively faster than the state-of-the-art methods.

        In general, ROL$_{\text{HC}}$ is more scalable than RL-based algorithms.
        For example, ROL$_{\text{HC}}$ could easily be applied on graphs with less than 200 variables with the computational power of a standard PC.
        ROL$_{\text{VI}}$ which uses value-iteration, is limited to small size graphs with at most 35-40 variables.
        This can be improved by using approximation RL algorithms such as vanilla policy gradient.
        Our ROL$_{\text{PG}}$ uses vanilla policy gradient.
        
        In terms of the complexity of our algorithms, herein, we discuss the number of CI tests required by our algorithm. 
        As we described in Appendix C.2, we used an algorithm for \textit{FindNeighbors} that requires $|Mb(X)|$ number of CI tests.
        Hence, Algorithm \ref{alg: compute cost} requires at most $(b-a)\alpha$ CI tests, where $\alpha$ is the size of the largest Markov boundary.
        Then, each iteration of Algorithm \ref{alg: compute cost} (the implementation of ROL$_{\text{HC}}$) performs at most $\mathcal{O}(n\ \text{maxSwap}^2  \alpha)$ CI tests.
        Hence, ROL$_{\text{HC}}$ performs at most $O(n \ \textit{maxIter}\ \textit{maxSwap}^2 \alpha)$ CI tests.

    \subsection{C.5 \quad Real-world Structures} 
        In Table \ref{tab: structure details}, we present more details about the real-world structures used in our experiments.
        In this table, $n$, $e$, $\Delta_{\text{in}}$, and $\Delta$ denote the number of vertices, number of edges, maximum in-degree, and maximum degree of the structures, respectively.
        
        \begin{table}[ht]
            \centering
            \begin{tabular}{N M{2cm}||M{1.5cm} M{1.5cm} M{1cm} M{1cm} M{1cm} M{1cm}}
                & Graph name & $n$ & $e$ &$\Delta_{\text{in}}$ &$\Delta$ \\
                \hline
                & Asia & 8 & 8 & 2 & 4\\
                & Sachs & 11 & 17 & 3 & 7\\
                & Insurance & 27 & 51 & 3 & 9\\
                & Water & 32 & 66 & 5 & 8\\
                & Alarm & 37 & 46 & 4 & 6\\
                & Ecoli70 & 46 & 70 & 4 & 11\\
                & Barley & 48 & 84 & 4 & 8\\
                & Hailfinder & 56 & 66 & 4 & 17\\
                & Carpo & 61 & 74 & 5 & 12\\
                & Hepar2 & 70 & 123 & 6 & 19\\
                & Arth150 & 107 & 150 & 6 & 20\\
            \end{tabular}
            \caption{Detailed information of the real-world structures used in the experiments.}
            \label{tab: structure details}
        \end{table}